\documentclass[10pt]{article} 
\usepackage[accepted]{tmlr}

\usepackage{amsmath,amssymb,amsfonts}
\usepackage[dvipsnames,table]{xcolor}
\usepackage{graphicx}
\usepackage{hyperref}
\usepackage[capitalize,nameinlink,noabbrev]{cleveref}
\usepackage{makecell}
\usepackage{multirow} 
\usepackage{colortbl}
\usepackage{diagbox}

\definecolor{lightred}{rgb}{1,0.8,0.8}
\definecolor{lightgreen}{rgb}{0.8,1,0.8}
\definecolor{linen}{rgb}{0.980,0.941,0.902}
\definecolor{nyu}{RGB}{87, 6, 140}
\definecolor{darkblue}{rgb}{0, 0, 0.5}
\hypersetup{colorlinks=true, citecolor=darkblue, linkcolor=darkblue, urlcolor=darkblue}

\usepackage{mathtools}
\usepackage{bbm}
\usepackage{booktabs} 
\usepackage{xspace}
\DeclarePairedDelimiterX{\infdivx}[2]{(}{)}{%
  #1\;\delimsize\|\;#2%
}

\newcommand{\calD}{\mathcal{D}}
\newcommand{\calN}{\mathcal{N}}
\newcommand{\calF}{\mathcal{F}}
\newcommand{\calQ}{\mathcal{Q}}

\newcommand{\DKL}{\mathbb{D}_{\text{KL}}\infdivx}
\newcommand\defines{\,\dot{=}\,}
\newcommand{\vbar}{\,|\,}
\newcommand{\kld}{KL divergence\xspace}

\usepackage{include/titletoc}
\usepackage{hyperref}
\usepackage{url}
\usepackage{appendix}
\usepackage{wrapfig}
\usepackage{colortbl}
\usepackage{tcolorbox}
\definecolor{lightred}{rgb}{1,0.8,0.8}
\definecolor{lightgreen}{rgb}{0.8,1,0.8}
\definecolor{bluu}{HTML}{0077b6}
\definecolor{cool}{HTML}{f72585}

\title{Data-Driven Priors for Uncertainty-Aware Risk Prediction of Clinical Deterioration using Multimodal Data}

\author{\name L. Julián Lechuga López \email leopoldo.lechuga@nyu.edu \\
      \addr New York University, New York University Abu Dhabi
      \AND
      \name Tim G. J. Rudner \email tim.rudner@utoronto.ca \\
      \addr University of Toronto
      \AND
      \name Farah E. Shamout \email farah.shamout@nyu.edu\\
      \addr New York University Abu Dhabi
      }

\def\month{09}  
\def\year{2026} 
\def\openreview{\url{https://openreview.net/forum?id=L6PmexvcFY}} 

\begin{document}

\maketitle

\begin{abstract}
Safe predictions are a crucial requirement for integrating predictive models into clinical decision support systems. 
One approach to improving trustworthiness is to enable models to express uncertainty about individual predictions.
However, current machine learning models frequently lack reliable uncertainty estimation, hindering real-world deployment. 
This limitation is particularly evident in multimodal settings, where models must effectively integrate heterogeneous information.
In this work, we propose \texttt{MedCertAIn}, a predictive uncertainty framework that leverages multimodal clinical data to improve model performance and reliability for in-hospital mortality risk prediction as an indicator of patient deterioration.
We design data-driven priors over neural network parameters using a hybrid strategy that considers cross-modal similarity in self-supervised latent representations and modality-specific data corruptions.
We train and evaluate the models with such priors using clinical time-series and chest X-ray images from the publicly available datasets MIMIC-IV and MIMIC-CXR. 
Our results show that \texttt{MedCertAIn} achieves competitive predictive performance and substantial gains in selective prediction compared with the evaluated deterministic and stochastic baselines.
These findings highlight the promise of data-driven priors in advancing robust, uncertainty-aware AI tools for high-stakes clinical applications.
Our implementation is publicly available at:
\href{https://github.com/jlaitue/medcertain}{jlaitue/medcertain}.
\end{abstract}

\section{Introduction}
\label{sec:introduction}

Considering the safety-critical nature of clinical practice, trustworthy implementation of machine learning models in healthcare requires robust uncertainty measurements \citep{begoli2019, gruber2023sources}.
Uncertainty Quantification (UQ) methods provide an essential layer of reliability in addition to accurate machine learning models for improved and safer decision-making \citep{nemani2023uncertainty}.
Uncertainty in machine learning models can be due to model parameters, noise and bias of the calibration data, or deployment of the model in an out-of-distribution scenario \citep{nado2022uncertainty}.
All of these are inherent characteristics of any real-world clinical setting.
Artificial Intelligence (AI) systems that reliably provide uncertainty measurements can enhance the confidence of healthcare professionals when using these systems and, at the same time, improve the effectiveness of the predictions \citep{seoni2023application}.

Nonetheless, the study of UQ within the area of machine learning for healthcare has been limited \citep{kompa2021second, uq_survey_lechuga}.
A comprehensive understanding of the uncertainty present in healthcare applications and how to address it is still fragmented \citep{han2011varieties}, partially due to the limited underlying theory on how to best adapt predictive uncertainty for clinical tasks \citep{begoli2019}. 
Other reasons include the complexity of scaling UQ in real-time clinical systems, limited empirical evaluation of different methods due to the lack of well-constructed priors by medical experts \citep{kurz2022uncertainty}, and the high prevalence of data shifts in real-world clinical applications.
All of these can negatively affect predictive performance \citep{ovadia2019can, xia2022benchmarking}, further emphasizing the need for better measurements of uncertainty in predictive models.

In addition, developing machine learning systems that are suitable for healthcare applications requires the use of different data modalities simultaneously to reflect the multimodal nature of the human decision-making process \citep{saab2024capabilities}.
However, existing work on UQ in healthcare has been mainly studied in the unimodal setting, with a particular focus on medical imaging applications \citep{gawlikowski2021survey}.
Hence, effective quantification of predictive uncertainty in the context of multimodal clinical problems remains a challenging and unsolved task \citep{Plex22}.

\begin{figure}[t!]
\centering
  \includegraphics[width=\linewidth]{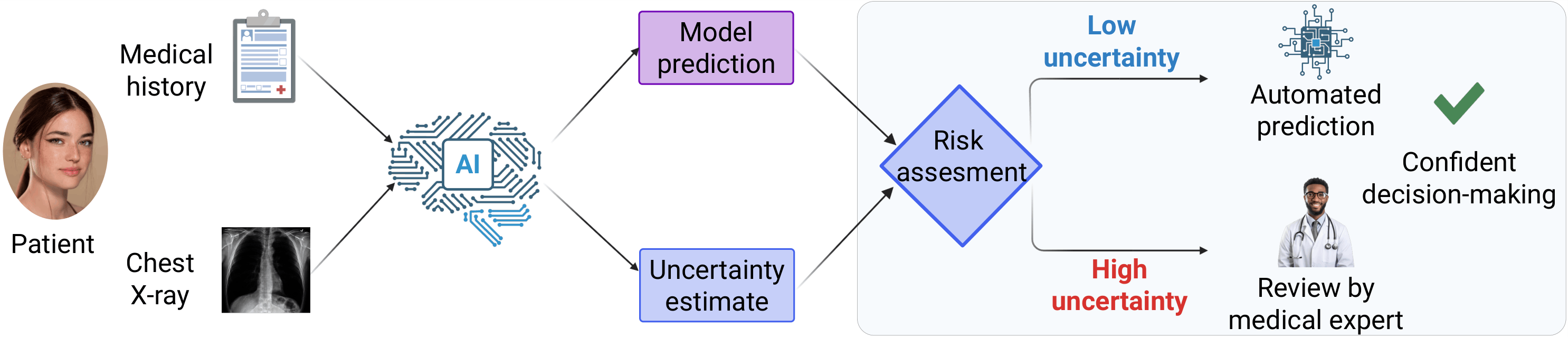}
    \caption{\small
    \textbf{Uncertainty-Aware Clinical Decision-Making.}
    Multimodal patient data are processed by an uncertainty-aware model to produce a prediction and corresponding uncertainty estimate. Clinicians can accept confident predictions while deferring uncertain cases for further review, improving workflow efficiency and supporting reliable clinical decision-making.
    }
  \label{fig:figure_1}
\end{figure}

To address these gaps, we introduce \texttt{MedCertAIn}, a multimodal uncertainty-aware framework for deterioration risk prediction, operationalized in this work as in-hospital mortality prediction.
The framework estimates predictive uncertainty and supports uncertainty-based ranking and deferral under selective prediction, providing a foundation for future human-review workflows (\Cref{fig:figure_1}).
Our key contributions are:
\begin{enumerate}
    \item We introduce \texttt{MedCertAIn}, an uncertainty-aware multimodal learning framework that incorporates data-driven priors into Bayesian neural networks trained using variational inference.
    The framework modifies the training objective without requiring changes to the underlying multimodal fusion architecture and supports uncertainty-based deferral for in-hospital mortality prediction.

    \item We propose a label-free pipeline for constructing multimodal data-driven priors by combining modality-specific input corruptions with low cross-modal similarity in self-supervised latent representations as a proxy for multimodal disagreement.
    This approach does not depend on expert-annotated examples of high-uncertainty clinical cases.

    \item We evaluate \texttt{MedCertAIn} using paired EHR time-series and chest X-ray data from MIMIC-IV and MIMIC-CXR.
    Across multiple mortality prediction horizons, modality configurations, and patient subgroups, \texttt{MedCertAIn} achieves competitive predictive performance and substantial gains in selective prediction, particularly in the multimodal setting, relative to deterministic models and alternative uncertainty-estimation methods.
    
\end{enumerate}

The paper is organized as follows: \Cref{sec:related_work} provides an overview of related work on multimodal learning and UQ in healthcare.
\Cref{sec:method} presents the proposed \texttt{MedCertAIn} framework and training methodology. 
\Cref{sec:experiments} describes the experimental setting and \Cref{sec:results} presents the results and empirical findings. 
\Cref{sec:discussion} discusses our contribution, its limitations, and future directions.

\section{Related Work}
\label{sec:related_work}

\subsection{Multimodal Learning in Healthcare}

Multimodal learning approaches seek to leverage complementary information from different data modalities to enhance the models' predictive capabilities  \citep{medfuse}. 
In healthcare, this paradigm is particularly relevant due to the heterogeneous nature of clinical data, which often includes structured electronic health records, medical imaging, physiological signals, and free-text reports.
Many approaches for combining information across modalities exist, with multimodal fusion being the most widely adopted strategy \citep{huang2020fusion, medfuse}.

Early work primarily focused on integrating multiple imaging modalities, or views, for tasks such as segmentation and disease characterization, particularly in neuroimaging \citep{zhang2020advances, calhoun2016multimodal}. More recent studies have extended multimodal learning to combine fundamentally different data types, such as imaging with time-series or tabular clinical data, demonstrating improved performance in prognostic and risk prediction tasks \citep{muhammad2021comprehensive}.
This trend is exemplified by multimodal systems developed for patient deterioration and outcome prediction, including COVID-19 prognosis using combined imaging and clinical variables \citep{shamout2021artificial, jiao2021prognostication}.
While multimodal integration enables richer patient representations, it also introduces additional sources of uncertainty arising from heterogeneity across modalities, imperfect alignment in time and semantics, and potential disagreement between modality-specific signals \citep{OOD_medical}.
These challenges are inherent in real-world scenarios, motivating the need for modeling approaches that explicitly account for uncertainty in multimodal settings \citep{kurz2022uncertainty}.

\subsection{Uncertainty Quantification in Medical Applications}

The safety-critical nature of clinical decision-making has motivated growing interest in UQ for machine learning models in healthcare \citep{gawlikowski2021survey, seoni2023application}.
Most existing work on UQ in medical applications has focused on unimodal settings, particularly medical imaging, where uncertainty estimates have been used to support tasks such as brain tumor segmentation \citep{jungo2018towards}, skin lesion classification \citep{devries2018leveraging}, and diabetic retinopathy detection \citep{filos2019systematic,Band2021benchmarking,nado2022uncertainty}.
These studies demonstrate the potential of UQ to improve interpretability and model trust by distinguishing confident predictions from ambiguous cases.

However, uncertainty in real-world healthcare systems arises from multiple sources, including data noise, dataset shift, model misspecification, and deployment in out-of-distribution clinical environments \citep{uq_survey_lechuga}.
Existing UQ approaches rarely account for these factors jointly or explicitly address the challenges introduced by multimodal clinical data. 
These limitations are particularly relevant to selective prediction and deferral, where uncertainty may arise from modality disagreement or missing and corrupted inputs \citep{foundationsmultimodal}.

\subsection{Stochastic Neural Networks and Variational Inference}
Stochastic neural networks extend deterministic neural networks by treating their parameters as random variables rather than fixed point estimates. This probabilistic formulation enables the model to capture epistemic uncertainty and provides a principled foundation for Bayesian inference over network weights.
Hence, for a given neural network $f(\cdot)$, a stochastic version $f(\cdot \,; \Theta)$ is defined in terms of stochastic parameters $\Theta$. 
For an observation model $p_{Y \mid X, \Theta}$ and a prior distribution over parameters $p_{\Theta}$, Bayesian inference provides a principled framework for modeling predictive uncertainty by inferring the posterior distribution over parameters given the observed data, $p_{\Theta \mid \mathcal{D}}$ \citep{mckay1992practical, neal1996bayesian}.
However, since neural networks are non-linear in their parameters, exact inference over the stochastic network parameters is analytically intractable.
Variational inference is an approach that seeks to avoid this intractability by framing posterior inference as finding an approximation $q_{\Theta}$ to the posterior $p_{\Theta | \calD}$ via the variational optimization problem:
\begin{align*}
    \min\nolimits_{q_{\Theta} \in \calQ_{\Theta}} \DKL{q_{\Theta}}{p_{\Theta | \calD}}
    \Longleftrightarrow
    \max\nolimits_{q_{\Theta} \in \calQ_{\Theta}}  \calF(q_{\Theta})
\end{align*}
where $\calF(q_{\Theta})$ is the variational objective:
\begin{align*}
    \calF(q_{\Theta})
    \hspace*{-1pt}\defines\hspace*{-1pt}
    \mathbb{E}_{q_{\Theta}}[\log p(y_{\calD} \vbar x_{\calD}, \Theta) ] \hspace*{-1pt}-\hspace*{-1pt} \DKL{q_{\Theta}}{p_{\Theta}} ,
\end{align*}
$\calQ_{\Theta}$ is a variational family of distributions~\citep{wainwright2008vi}, $\mathbb{D}_{\textrm{KL}}$ is the Kullback-Leibler divergence \citep{kullback1951information}, and $(x_{\calD}, y_{\calD})$ are the training data.
One particularly simple type of variational inference is Gaussian mean-field variational inference \citep{blundell2015mfvi,graves2011practical}, where the posterior distribution over network parameters is approximated by a Gaussian distribution with a diagonal covariance matrix.
This method enables stochastic optimization and can be scaled to large neural networks~\citep{hoffman2013svi}.
However, Gaussian mean-field variational inference has been shown to underperform deterministic neural networks when uninformative, standard Gaussian priors are used~\citep{ovadia2019can, fsvi}. 
\begin{figure*}[t!]
\centering
\includegraphics[width=\textwidth]{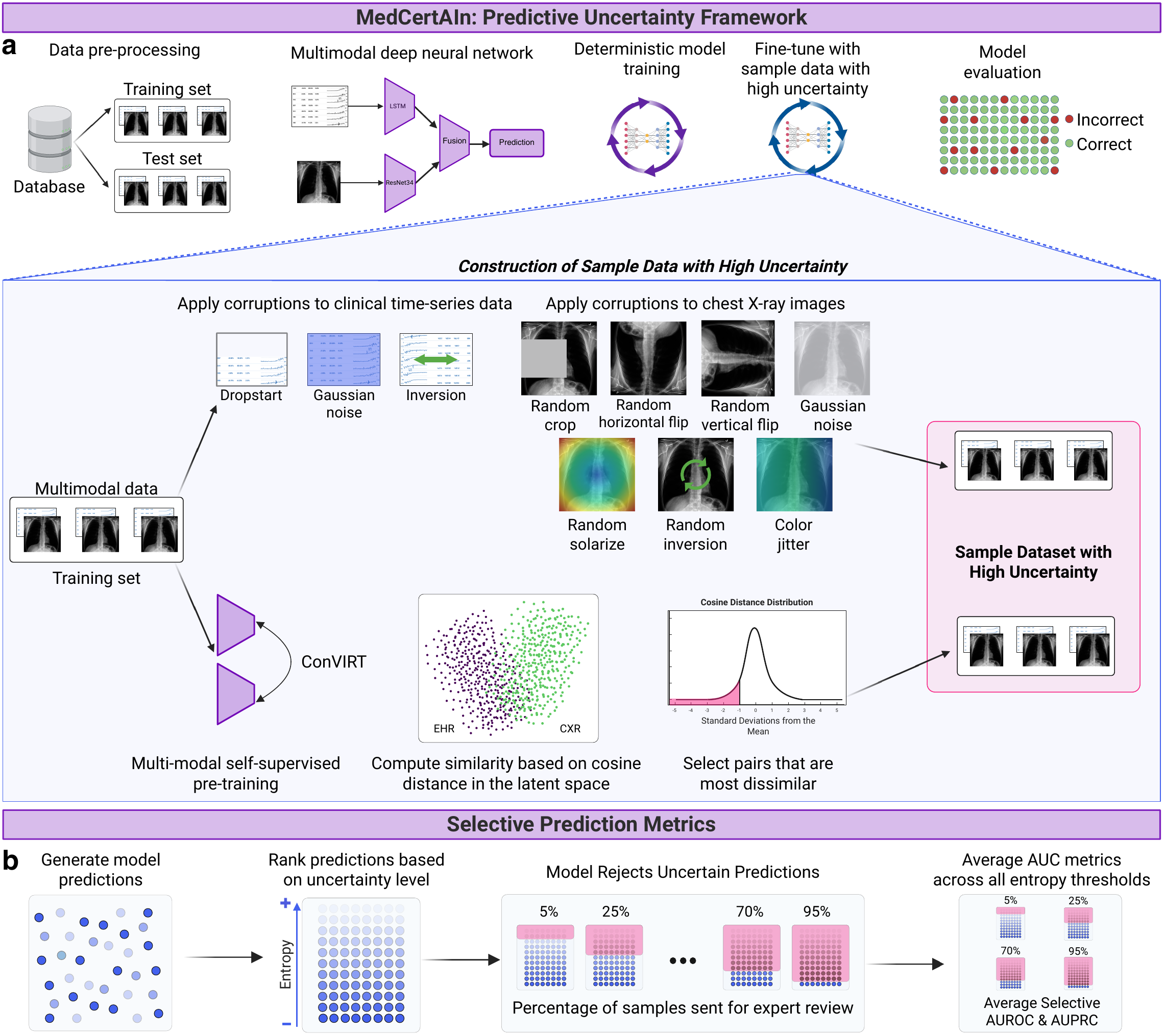}
    \caption{\small
    \textbf{\texttt{MedCertAIn}: A Predictive Uncertainty Framework.}
    \textbf{a)} We construct paired multimodal samples for training and evaluation and generate high-uncertainty context data. From the training set, we create corrupted pairs using noise and transformations and pretrain a self-supervised model to identify low-similarity pairs in latent space, representing high cross-modal uncertainty. We combine these context sets to define a data-driven prior, fine-tune the model using Bayesian learning, and evaluate standard and selective performance.
    \textbf{b)} \textbf{Selective Prediction Metrics.} We rank predictions by Shannon entropy and evaluate performance on retained subsets across coverage thresholds (0--100\%). Averaging across thresholds yields selective AUROC and AUPRC, while the coverage curve supports selection of an operating point for deferring the least reliable predictions.
    }
  \label{fig:figure_2}
\end{figure*}

\section{Method}
\label{sec:method}

\texttt{MedCertAIn} is a multimodal Bayesian learning framework for predictive uncertainty estimation and selective prediction.
It constructs data-driven priors over neural network parameters using context samples selected to represent regions of high uncertainty in the multimodal input space.
These priors are incorporated into a variational training objective that encourages the model to express greater uncertainty on the selected context samples while optimizing the primary prediction task.
Because \texttt{MedCertAIn} modifies the training objective rather than the multimodal fusion architecture, it can be applied to existing fusion models without altering their underlying design.

\subsection{Deriving the Optimization Objective}

We consider a supervised multimodal fusion task on a dataset
$\mathcal{D} \doteq \{(x^{ehr}_n, x^{cxr}_n, y_n)\}_{n=1}^{N}
= (X^{ehr}_{\mathcal{D}}, X^{cxr}_{\mathcal{D}}, Y_{\mathcal{D}})$,
where each sample consists of paired electronic health record time-series and chest X-ray images.
As illustrated in \Cref{fig:figure_2}.a, the two modalities are processed by modality-specific encoders $\Phi^{ehr}$ and $\Phi^{cxr}$, respectively.
The resulting feature representations are concatenated and passed to a classifier $g(\cdot)$ to produce a fusion prediction $\hat{y}$, which is optimized using supervised learning with respect to the ground truth labels $y \in \{0,1\}$.

To mitigate the limitations of variational inference under non-informative priors, we use information from our two input modalities to construct data-driven priors that favor parameter values inducing reliable uncertainty-aware predictions \citep{rudner2024mind}.
We consider a standard parametric observation model $p_{Y | X, \Theta}(y \vbar x, \theta; f)$, and let the mapping $f$ be defined by \mbox{$f(\cdot \,; \theta) \defines h(\cdot \,; \theta_{h}) \theta_{L}$}, where $h(\cdot \,; \theta_{h})$ is the post-activation output of the penultimate layer, $\Theta_{L}$ is the set of stochastic final-layer parameters, $\Theta_{h}$ is the set of stochastic non-final-layer parameters, and $\Theta \defines \{ \Theta_{h} , \Theta_{L}\}$ is the full set of stochastic parameters.
In preparation for constructing a data-driven prior, we denote a batch of informative context inputs with corresponding context labels by
$x_c=\{x_1,\ldots,x_M\}, y_c=\{y_1,\ldots,y_M\},$ respectively, and let \(p_{X_c,Y_c}\) be a joint distribution over context batches.
The context inputs are chosen to represent a priori high-uncertainty regions, such as corrupted or cross-modally ambiguous samples.

To construct an uncertainty-aware data-driven prior, we consider a Bernoulli random variable $\check{Z}$, denoting whether a given set of neural network parameters induces predictions that exhibit some desired property (e.g., high uncertainty on a set of evaluation points).
Furthermore, we define a {\em context observation model}
$\check{p}_{\smash{\check{Z} \vbar \Theta}}(\check{z} \vbar \theta ; f, p_{\smash{{X}_{c}, {Y}_{c}}})$---which denotes the likelihood of observing an auxiliary outcome \(\check{z}\) under $\check{p}_{\smash{\check{Z} \vbar \Theta}}$, given $\theta$, and $p_{\smash{{X}_{c}, {Y}_{c}}}$---and specify a {\em base prior} over the model parameters, $p_{\Theta}(\theta)$.
Here, the event $\check{z}=1$ corresponds to the parameter setting inducing the desired uncertainty behavior on the context points.
Using the context data and the above context observation model, we can define a data-driven prior $p(\theta \vbar \check{z} ; p_{\smash{{X}_{c}, {Y}_{c}}})$, conditioned on the desired event occurring.

Unfortunately, since the full distribution $p(\theta \vbar \check{z} ; p_{\smash{{X}_{c}, {Y}_{c}}})$ is intractable for all but the simplest context observation models, we cannot use this prior directly.
Nevertheless, \Cref{appsec:data_driven_priors} shows how to obtain a tractable optimization objective without computing the full data-driven prior density and instead only requires specifying an auxiliary context cost function, $c(X_c,Y_c,\Theta)$.
This cost function defines the context observation model and is chosen to penalize parameter values $\theta$ that induce predictions for which the expected cost under $p_{\smash{{X}_{c}, {Y}_{c}}}$ is high.
For training our stochastic model, we therefore extend the variational objective with an uncertainty regularization term that makes use of the set of context points:
\begin{align*}
\begin{split}
    \calF(q_{\Theta})
    \defines
    \underbrace{\mathbb{E}_{q_{\Theta}} [ \log p(y_{\calD} \vbar x_{\calD} , \Theta ; f) ]}_{\textrm{Expected log-likelihood}} 
    - 
    \underbrace{\DKL{q_{\Theta}}{p_{\Theta}}}_{\textrm{KL regularization}}
    - \underbrace{ \mathbb{E}_{q_{\Theta}} [ \mathbb{E}_{p_{{X}_{c}, {Y}_{c}}} [c(\smash{{X}_{c}, {Y}_{c}} , \Theta) ]]
    }
    _{\textrm{Uncertainty regularization}}.
\end{split}
\end{align*}
Letting $p_{{Y}_{c} | {X}_{c}}({y}_{c} \vbar {x}_{c}) = \delta(\mathbf{0}-y_c)$ to encourage high uncertainty in the predictions on the set of context points, where $\delta(\cdot)$ is the Dirac delta function, we obtain the simplified objective:
\begin{align*}
\begin{split}
    \calF(q_{\Theta})
    \defines
    \underbrace{\mathbb{E}_{q_{\Theta}} [ \log p(y_{\calD} \vbar x_{\calD} , \Theta ; f) ]}_{\textrm{Expected log-likelihood}} 
    &- \underbrace{\DKL{q_{\Theta}}{p_{\Theta}}}_{\textrm{KL regularization}}
    - \underbrace{ \mathbb{E}_{q_{\Theta}} [ \mathbb{E}_{p_{{X}_{c}}} [ c({X}_{c}, \smash{\mathbf{0}} , \Theta) ]]
    }
    _{\textrm{Uncertainty regularization}}.
\end{split}
\end{align*}

In practice, each training step performs a forward pass on the combined training and context sets.
We optimize the expected log-likelihood using binary cross-entropy on the supervised training labels, and compute the KL and uncertainty regularization terms using Monte Carlo estimates under \(q_\Theta\).
Gradients are obtained using the reparameterization trick for mean-field variational inference \citep{blundell2015mfvi}.
The complete derivation of the tractable objective under the proposed uncertainty-aware prior is provided in \Cref{sec:variational_objective}.

\subsection{Constructing the Context Set in A Priori High-Uncertainty Regions}

We define a context set $({X}_{c},{Y}_{c})$ drawn from a distribution $p_{{X}_{c},{Y}_{c}}$, representing samples in a priori high-uncertainty regions (i.e., distributionally shifted or ambiguous points). 
From the multimodal training set  $(X^{ehr}_{\mathcal{D}}, X^{cxr}_{\mathcal{D}})$, we construct a context set $({X}^{ehr}_{\mathcal{C}}, {X}^{cxr}_{\mathcal{C}})$ using two label-free strategies which enable data-driven prior construction with minimal human intervention: (1) applying controlled corruptions to both modalities, and (2) selecting samples with low cross-modal similarity in a self-supervised latent space. 
The latter is motivated by the assumption that modality mismatch corresponds to higher predictive uncertainty. 

\subsubsection{Modality-specific Data Perturbations}
We induce a controlled distribution shift by applying data corruptions to the original training set. 
For EHR time-series, we construct ${X}^{ehr}_{\mathcal{C}}$ by (i) dropping an initial segment of the sequence up to a threshold, (ii) adding Gaussian noise, and (iii) reversing the time dimension. 
For chest X-rays, we construct ${X}^{cxr}_{\mathcal{C}}$ using seven perturbations: random crop, horizontal/vertical flip, Gaussian blur, solarize, invert, and color jitter. 
Corruption magnitudes are fixed during training using commonly adopted augmentation settings.

\subsubsection{Cross-Modal Similarity in Multimodal Latent Space}
Extending previous work \citep{lopez2023informativepriorsimprovereliability}, we also construct a context set by selecting samples whose modality representations are dissimilar in a self-supervised latent space. 
We obtain per-modality representations $(\Phi^{\textrm{ehr}},\Phi^{\textrm{cxr}})$ using a ConVIRT model trained with the InfoNCE loss \citep{zhang2022contrastive,medfuse}. 
We compute cosine similarity scores and select samples falling in the left tail of the similarity distribution.

\paragraph{Inter-modal similarity.}
For each training sample, we compute the cross-modal cosine similarity and define the selection threshold $t_1$:
\begin{align*}
    t_1 &= \bar{\alpha} - v\cdot\sigma_\alpha,
    \qquad
    \alpha_i =\cos(\Phi^{\textrm{ehr}}_i, \Phi^{\textrm{cxr}}_i), 
    \qquad
    \bar{\alpha} = \frac{1}{n}\sum_{i=1}^n \alpha_i,
    \qquad
    \sigma_{\alpha} = \operatorname{std}\{\alpha_i\}_{i=1}^n,
    \qquad
    v\in[1,2].
\end{align*}

We define the selected index set as $\mathcal{C}_1 = \{i\in\{1,\dots,n\}: \alpha_i < t_1\}$, and the corresponding context sets:
\begin{align*}
    X^{\textrm{ehr}}_{\mathcal{C}_1}
    = \{X^{\textrm{ehr}}_i : i\in\mathcal{C}_1\}, 
    \qquad
    X^{\textrm{cxr}}_{\mathcal{C}_1}
    = \{X^{\textrm{cxr}}_i : i\in\mathcal{C}_1\}.
\end{align*}

\paragraph{Inter- and intra-modal similarity.}

We additionally incorporate intra-modal similarity between each training sample
and the mean latent representation vector for each modality:
\begin{align*}
   \beta_i &= \cos\left(\Phi^{\textrm{ehr}}_i, \overline{\Phi}^{\textrm{ehr}}\right),
   &
   \gamma_i &= \cos\left(\Phi^{\textrm{cxr}}_i, \overline{\Phi}^{\textrm{cxr}}\right),
   \qquad
   \overline{\Phi}^{\textrm{ehr}} = \frac{1}{n}\sum_{i=1}^n \Phi_i^{\textrm{ehr}},
    \qquad
    \overline{\Phi}^{\textrm{cxr}} = \frac{1}{n}\sum_{i=1}^n \Phi_i^{\textrm{cxr}}.
\end{align*}
We aggregate the similarities by simple averaging and define the selection threshold $t_2$:
\begin{align*}
    t_2 &= \bar{\delta} - v\sigma_{\delta},
    \qquad
    \delta_i = \frac{\alpha_i + \beta_i + \gamma_i}{3}, 
    \qquad
    \bar{\delta} = \frac{1}{n}\sum_{i=1}^n \delta_i,
    \qquad
    \sigma_{\delta} = \operatorname{std}\{\delta_i\}_{i=1}^n.
\end{align*}
We define the selected index set as $\mathcal{C}_2 = \{i\in\{1,\dots,n\}: \delta_i < t_2\}$, and the corresponding context sets:
\begin{align*}
    X^{\textrm{ehr}}_{\mathcal{C}_2}
    = \{X^{\textrm{ehr}}_i : i\in\mathcal{C}_2\},
    \qquad
    X^{\textrm{cxr}}_{\mathcal{C}_2}
    = \{X^{\textrm{cxr}}_i : i\in\mathcal{C}_2\}.
\end{align*}

Finally, for training \texttt{MedCertAIn}, the high-uncertainty sample set is
constructed by combining the corrupted training examples with either the
inter-modal samples \(\mathcal{C}_1\) or the inter- and intra-modal samples
\(\mathcal{C}_2\) selected using self-supervised latent-space similarity.

\subsection{Uncertainty Measurement}

To quantify predictive uncertainty, we compute uncertainty scores individually for each test sample based on the predictive distribution of the model. 
For a given input $x$, the stochastic network induces a predictive distribution $p(y\mid x)$, obtained via Monte Carlo sampling from the variational posterior.
We quantify uncertainty using the Shannon entropy of the predictive distribution, where higher entropy indicates greater predictive uncertainty:
\begin{equation*}
    \mathcal{H}(p(y \mid x)) 
    = - \sum_{c \in \{ 0, 1\}} p(y = c \mid x) 
    \log p(y = c \mid x).
\end{equation*}

Shannon entropy provides a simple and widely used proxy for predictive uncertainty in Bayesian neural
networks \citep{rudner2023fseb, rudner2023fsmap}. 
In our setting, entropy scores are computed for each sample and used to rank predictions by uncertainty, thereby identifying potential candidates for deferral under selective prediction.

\section{Experimental Setting}
\label{sec:experiments}

\setlength{\tabcolsep}{4pt}
\begin{wraptable}{r}{0.5\textwidth}
\vspace{-22pt}
\caption{\textbf{Dataset Summary}: Summary of multimodal data samples used during training for the in-hospital mortality task. 
Only our proposed method \texttt{MedCertAIn} makes use of the high-uncertainty set (5,347 samples) which is constructed by combining corrupted training examples with 862 additional samples selected via cross-modal similarity in the self-supervised latent space.}
    \vspace{2mm}
    \small
    \centering
    \begin{tabular}{lcc}
    \toprule
    \diagbox[width=9em]{\textbf{Data split}}{\textbf{Model}} & 
    \textbf{Baseline Models} & \texttt{MedCertAIn} \\
    \midrule
   Training & 4,485 & 4,485 \\
   Validation & 488 & 488 \\
   Test & 1,242 & 1,242 \\ \midrule
   High Uncertainty Set & -- & 5,347 \\
    \bottomrule
    \end{tabular}
\label{table:table_data_medcertain}
\vspace{-10pt}
\end{wraptable}

\subsection{Dataset}
To train and evaluate our framework, we used two publicly available modalities: EHR time-series data from MIMIC-IV \citep{mimiciv} and chest X-ray images from MIMIC-CXR \citep{mimiccxr}.
MIMIC-IV contains more than 60,000 ICU stays collected between 2001 and 2019.
We only used complete paired samples, such that each example includes both modalities with no missing data.
We focus on in-hospital mortality prediction, a clinically relevant task for ICU decision support systems \citep{sadeghi2018early, awad2017early}, to support clinical decision-making and resource allocation by identifying patients at higher risk of deterioration, enabling timely intervention \citep{rajpurkar2022ai}.

Mortality prediction is a binary classification task that uses data from the first 48 hours of the ICU stay to predict in-hospital mortality \citep{medfuse}.
We excluded ICU stays shorter than 48 hours, resulting in 6,215 paired patient samples.
We first held out 20\% of the data for testing and then split the remaining 80\% into training and validation sets, yielding approximate overall proportions of 70\%, 10\%, and 20\%, respectively. 
All splits were disjoint at the patient level.
We follow \citet{medfuse} to pre-process and align the clinical time-series data and chest X-ray images for this task. \Cref{table:table_data_medcertain} summarizes the multimodal sample sizes across training, validation, test and context set splits used for \texttt{MedCertAIn} and \Cref{table:dataset_sizes} summarizes the modality-specific dataset sizes.

\subsection{Multimodal Backbone Neural Network}

We adopt MedFuse \citep{medfuse} as the backbone neural network architecture, a simple and robust fusion module for multimodal clinical prediction using paired EHR time-series and imaging data.
MedFuse follows an early-fusion design, where modality-specific encoders process modalities before their representations are combined for downstream prediction.
Given a paired multimodal sample $(x^{ehr}, x^{cxr}) \in \mathcal{D}$, $\Phi_{\textrm{ehr}}$ encodes EHR time-series using a two-layer long short-term memory (LSTM) network  \citep{lstm}, which captures temporal dependencies in sequential clinical measurements.
Chest X-ray images are encoded using a ResNet-34 convolutional neural network $\Phi_{\textrm{cxr}}$ \citep{resnet50}, extracting high-level spatial and semantic features from medical images.
Finally, modality-specific representations are concatenated and passed to a classifier $g(\cdot)$ with a sigmoid activation to produce the predicted mortality risk $\hat{y}$.

\subsection{Model Baselines}
We compare \texttt{MedCertAIn} against five primary baselines: the deterministic MedFuse backbone \citep{medfuse}, temperature scaling \citep{guo2017calibration}, deep ensembles \citep{lakshminarayanan2016simple}, MC Dropout \citep{gal2016dropoutbayesianapproximationrepresenting}, and Gaussian mean-field variational inference with an uninformative prior \citep{blundell2015mfvi,graves2011practical}.
These baselines allow us to compare \texttt{MedCertAIn} with its underlying deterministic architecture, established post-hoc calibration and uncertainty-estimation methods, and Bayesian learning without informative data-driven priors. 
All methods are evaluated using identical data splits and experimental protocols. 
In addition, \Cref{sec:deterministic_models_comparison} compares MedFuse with two recent deterministic multimodal architectures, DrFuse \citep{yao2024drfuse} and MeTra \citep{khader2023medical}, across all mortality prediction horizons, providing a broader reference for the performance of our chosen backbone relative to existing multimodal methods.

\subsection{Evaluation Metrics}
We evaluate test performance using AUROC and AUPRC. 
To assess predictive uncertainty, we additionally compute selective prediction metrics. 
Selective prediction introduces a ``reject option'' $\bot$ via a selection function $s:\mathcal{X}\to\mathbb{R}$ that determines whether to output a prediction for an input $x\in\mathcal{X}$ \citep{el2010foundations}. 
Given an entropy threshold $\tau$, predictions with entropy at most $\tau$ are retained, while those exceeding $\tau$ are rejected:
\begin{align*}
    p(y \mid x,\theta;f), & \text{ if } s(x)\leq\tau
    \begin{cases}
          p(y\,|\,x,\mathbf{\theta};f), & \text{if}\ s\le \tau \\
          \bot, & \text{otherwise.}
    \end{cases}
\end{align*}

We vary $\tau$ to obtain rejection rates from $0\%$ to $99\%$ and compute AUROC and AUPRC on each retained subset. 
We report the averages across rejection rates as selective AUROC and selective AUPRC (\Cref{fig:figure_2}.b).
Selective prediction metrics quantify predictive performance together with a model’s uncertainty, providing an intuitive measure of reliability. 
By sweeping the rejection range, we assess how effectively each method ranks predictions by uncertainty. 
A deployment-specific threshold would instead depend on the acceptable risk and available review capacity of the clinical setting.

\subsection{Hyperparameter Tuning and Model Selection}

\subsubsection{Deterministic models} For each task, we run 50 hyperparameter configurations, each evaluated over five runs with different random seeds (250 runs total). 
For every run, we randomly sample train/validation splits and initialize model weights. 
We sample the learning rate uniformly from $[10^{-5}, 10^{-2}]$, select the regularizer scale from $\{0, 0.1, 1, 10, 100\}$, and the number of training epochs from $\{5, 10, 15, 20, 30\}$. 
We use a fixed batch size of 16 and cosine decay with $\alpha=0$. 
For each run, we save the checkpoint from the final training epoch and select the configuration with the highest average validation AUROC across the five seeds.
Using the selected configuration, we retrain on the combined training and validation data and report test performance averaged over five random initializations. 
We compute standard errors across these five final runs.

\subsubsection{Stochastic models} We initialize stochastic models from the five best deterministic checkpoints and perform hyperparameter tuning using the same protocol (50 configurations $\times$ 5 seeds; 250 runs per task). 
The learning rate is sampled one order of magnitude above and below the best deterministic learning rate, and training epochs are sampled from $\{5, 10, 15, 20, 25, 30\}$. 
We use batch size 16, cosine decay with $\alpha=0$, and sample the context batch size from $\{16, 32\}$ due to the additional context data used for Bayesian learning. 
As stochastic models include mean and variance parameters and require forward passes over sampled context points, fine-tuning is more computationally expensive than the deterministic setting.
In Appendix \Cref{table:stochastic_hyperparams}, we summarize the stochastic search space.
With the optimal configuration, we train on the full training plus validation data and evaluate the stochastic models with five different random seed initializations. 
This procedure is repeated for each stochastic variant trained with different high-uncertainty context sets.

All models are trained with Adam \citep{adam_optim} and evaluated on the test set reporting means and standard errors over five seeds. 
Experiments were run on NVIDIA V100 GPUs.
Additional dataset, ConVIRT pretraining, stochastic hyperparameter search, and implementation details are provided in \Cref{app:experimental-details}.

\begin{figure*}[t!]
    \centering
    \includegraphics[width=\textwidth]{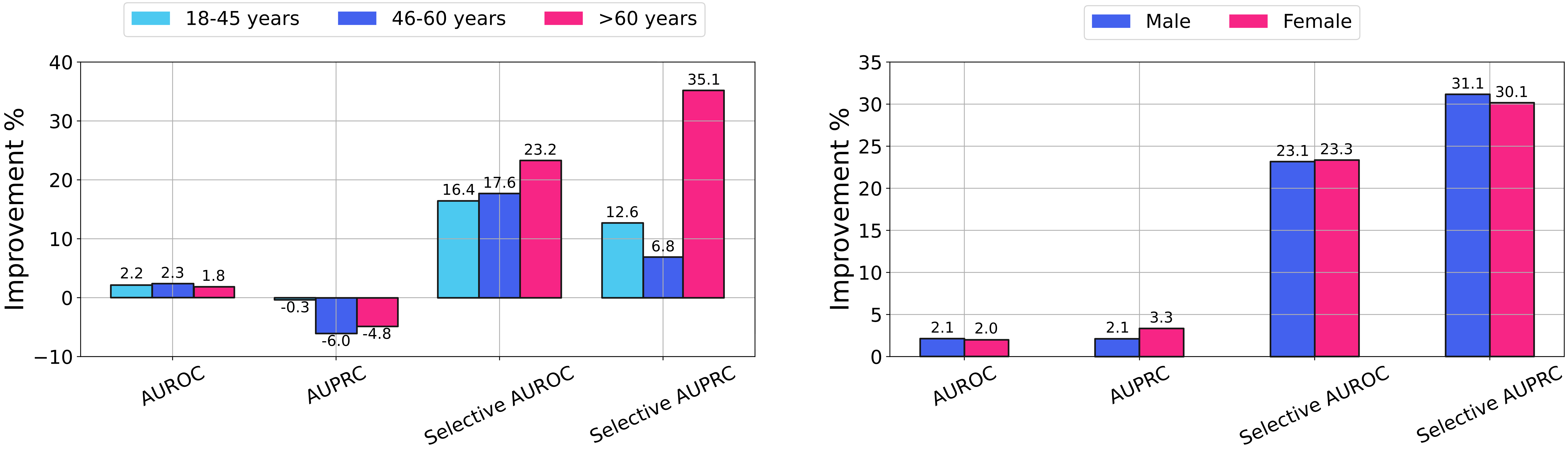}
    \vspace{-10pt}
        \caption{
        \small \textbf{\texttt{MedCertAIn} Achieves Strong Predictive Performance and Reliable Selective Prediction.}
        \textbf{a)} Standard AUROC and AUPRC curves for \texttt{MedCertAIn} and representative deterministic and stochastic baselines show relatively small differences in predictive performance.
        \textbf{b)} Larger differences emerge under selective prediction, where \texttt{MedCertAIn} achieves stronger and more stable performance across rejection thresholds and random seeds.
        \textbf{c)} Calibration and relative-improvement analyses further characterize the differences among the evaluated methods.
        \footnotesize{Shaded regions represent variation across five random seeds.}
        }
    \label{fig:figure_3}
\end{figure*}

\section{Results}
\label{sec:results}

\subsection{\texttt{MedCertAIn} Achieves Strong Predictive Performance and Reliable Selective Prediction}

\Cref{table:main_table} reports the main results for 48-hour in-hospital mortality prediction. 
We compare \texttt{MedCertAIn} with the deterministic MedFuse backbone \citep{medfuse}, temperature scaling \citep{guo2017calibration}, deep ensembles \citep{lakshminarayanan2016simple}, MC Dropout \citep{gal2016dropoutbayesianapproximationrepresenting}, and a Bayesian neural network with an uninformative Gaussian prior \citep{fsvi}.
Differences among the baselines are relatively small for standard predictive performance, with AUROC ranging from $0.819$ to $0.822$ and AUPRC ranging from $0.466$ to $0.485$. 
Nevertheless, \texttt{MedCertAIn} achieves the strongest performance, with an AUROC of $0.835\pm0.001$ and an AUPRC of $0.498\pm0.002$. 
Relative to the uninformative-prior BNN, which provides the most direct comparison for isolating the effect of our data-driven prior, these results correspond to improvements of $1.8\%$ in AUROC and $2.7\%$ in AUPRC.

Beyond standard predictive performance, selective prediction evaluates whether uncertainty estimates can identify clinical cases that should be deferred for further review. 
\texttt{MedCertAIn} achieves a selective AUROC of $0.857\pm0.005$ and a selective AUPRC of $0.599\pm0.002$, outperforming all evaluated baselines. 
Compared with the uninformative-prior BNN, this represents improvements of $20.5\%$ in selective AUROC and $67.8\%$ in selective AUPRC. 
The selective prediction curves in \Cref{fig:figure_3} further show that \texttt{MedCertAIn} maintains stronger and more stable performance across rejection thresholds and random seeds. 
These findings indicate that the gains do not arise merely from introducing stochastic Bayesian parameters: the proposed data-driven prior improves both predictive performance and the ranking of uncertain cases relative to an otherwise comparable BNN with an uninformative prior.
In \Cref{sec:extended_results}, we report complementary performance and calibration results for 3-month, 6-month, and 1-year mortality to assess whether these gains persist beyond the 48-hour prediction task.

\setlength{\tabcolsep}{8.5pt}
\begin{table}[t]
\caption{\textbf{\texttt{MedCertAIn} Achieves Strong Predictive Performance and Reliable Selective Prediction.} Performance on 48-hour in-hospital mortality prediction compared with deterministic, post-hoc calibration, ensemble, and stochastic uncertainty baselines. 
\texttt{MedCertAIn} achieves the strongest performance across all metrics, with particularly large improvements in selective AUROC and selective AUPRC relative to both MedFuse and the uninformative-prior BNN.}

\vspace{2mm}
    \small
    \centering
    \begin{tabular}{lcccc}
    
    \toprule
    
    \diagbox[width=9em]{\textbf{Model}}{\textbf{Metric}} & \textbf{AUROC} & \textbf{AUPRC} & \textbf{\makecell{Selective \\ AUROC}} & \textbf{\makecell{ Selective \\ AUPRC}}\\
    
     \midrule
   
   MedFuse \citep{medfuse} & 0.819\tiny{$\pm$0.000} & 0.466\tiny{$\pm$0.002} & 0.659\tiny{$\pm$0.005} & 0.203\tiny{$\pm$0.006} \\[2pt]

   Temp. Scaling \citep{guo2017calibration} & 0.820\tiny{$\pm$0.000} & 0.470\tiny{$\pm$0.002} & 0.657\tiny{$\pm$0.006} & 0.206\tiny{$\pm$0.011} \\[2pt]

   Deep Ensemble
   \citep{lakshminarayanan2016simple} & 
   0.819\tiny{$\pm$0.000} & 0.469\tiny{$\pm$0.000} & 0.657\tiny{$\pm$0.000} & 0.208\tiny{$\pm$0.000} \\[2pt]
    
   MC Dropout \citep{gal2016dropoutbayesianapproximationrepresenting} & \underline{0.822\tiny{$\pm$0.002}} & 0.480\tiny{$\pm$0.008} & 0.681\tiny{$\pm$0.007} & 0.233\tiny{$\pm$0.014} \\[2pt]

   Uninformative Prior \citep{fsvi} & 0.820\tiny{$\pm$0.004} & \underline{0.485\tiny{$\pm$0.006}} & 0.711\tiny{$\pm$0.034} & 0.357\tiny{$\pm$0.058} \\[2pt]

   \textbf{\texttt{MedCertAIn}} (\textbf{Ours}) & \cellcolor[gray]{0.85}\textbf{0.835\tiny{$\pm$0.001}} & \cellcolor[gray]{0.85}\textbf{0.498\tiny{$\pm$0.002}} & \cellcolor[gray]{0.85}\textbf{0.857\tiny{$\pm$0.005}} & \cellcolor[gray]{0.85}\textbf{0.599\tiny{$\pm$0.002}} \\
   
    \midrule
    Improvement vs MedFuse (\%) 
   & \cellcolor{lightgreen}+2.0\% 
   & \cellcolor{lightgreen}+6.9\% 
   & \cellcolor{lightgreen}+30.0\% 
   & \cellcolor{lightgreen}+195.1\% \\

   Improvement vs Uninformative Prior (\%) 
   & \cellcolor{lightgreen}+1.8\% 
   & \cellcolor{lightgreen}+2.7\% 
   & \cellcolor{lightgreen}+20.5\% 
   & \cellcolor{lightgreen}+67.8\% \\

    \bottomrule
    \end{tabular}
   \label{table:main_table}
    
\end{table}
\setlength{\tabcolsep}{14.2pt}
\begin{table}[t]
    \caption{\small \textbf{Comparison Across Multimodal Data-Driven Priors.} Performance on 48-hour in-hospital mortality prediction across alternative strategies for constructing the high-uncertainty context set. Combining latent-space similarity with random corruptions in \texttt{MedCertAIn I} and \texttt{MedCertAIn II} improves both predictive and selective performance, with \texttt{MedCertAIn I} achieving the strongest results across all metrics.}

    \vspace{2mm}
    \label{table:table_6}
    \small
    \centering
    \begin{tabular}{lcccc}
        \toprule
        \textbf{Ablation} & \textbf{AUROC} & \textbf{AUPRC} & \textbf{\makecell{Selective \\ AUROC}} & \textbf{\makecell{Selective \\ AUPRC}} \\
        \midrule
        
        Uninformative Prior & 0.820\tiny{$\pm$0.004} & 0.485\tiny{$\pm$0.006} & 0.711\tiny{$\pm$0.034} & 0.357\tiny{$\pm$0.058}\\
        Random Corruptions & 0.832\tiny{$\pm$0.001} & 0.494\tiny{$\pm$0.002} & 0.841\tiny{$\pm$0.006} & 0.579\tiny{$\pm$0.011}\\
        Inter-Modal Similarity & 0.821\tiny{$\pm$0.001} & 0.470\tiny{$\pm$0.002} & 0.749\tiny{$\pm$0.013} & 0.423\tiny{$\pm$0.032}\\
        Inter- and Intra-Modal Similarity & 0.821\tiny{$\pm$0.001} & 0.470\tiny{$\pm$0.002} & 0.755\tiny{$\pm$0.007} & 0.434\tiny{$\pm$0.020}\\

        \midrule

       \texttt{MedCertAIn I} & \cellcolor[gray]{0.85}\textbf{0.835\tiny{$\pm$0.001}} & \cellcolor[gray]{0.85}\textbf{0.498\tiny{$\pm$0.002}} & \cellcolor[gray]{0.85}\textbf{0.857\tiny{$\pm$0.005}} & \cellcolor[gray]{0.85}\textbf{0.599\tiny{$\pm$0.002}} \\
        \texttt{MedCertAIn II} & \underline{0.834\tiny{$\pm$0.001}} & \underline{0.497\tiny{$\pm$0.001}} & \underline{0.848\tiny{$\pm$0.008}} & \underline{0.590\tiny{$\pm$0.006}} \\
        \bottomrule
    \end{tabular}
\end{table}

\subsection{Comparison Across Multimodal Data-Driven Priors}

We ablate the strategies used to select high-uncertainty context samples to quantify how different constructions of the data-driven prior affect performance. 
\Cref{table:table_6} compares an uninformative Gaussian prior with priors constructed using random corruptions, inter-modal similarity, or combined inter- and intra-modal similarity. 
We additionally evaluate \texttt{MedCertAIn I}, which combines inter-modal similarity with corruptions, and \texttt{MedCertAIn II}, which combines inter- and intra-modal similarity with corruptions. 
\texttt{MedCertAIn I} is the primary variant reported throughout this study as it achieves the strongest overall performance.

Relative to the uninformative-prior baseline (AUROC/AUPRC: $0.820/0.485$; selective AUROC/AUPRC: $0.711/0.357$), random corruptions improve all four metrics to $0.832/0.494$ and $0.841/0.579$, respectively. The similarity-only variants maintain a similar AUROC ($0.821$) but reduce AUPRC ($0.470$), while moderately improving both selective metrics: inter-modal similarity achieves $0.749/0.423$, and combined inter- and intra-modal similarity achieves $0.755/0.434$. 

Combining similarity-based selection with corruptions yields the strongest results, with \texttt{MedCertAIn I} achieving $0.835/0.498$ on the standard metrics and $0.857/0.599$ on the selective metrics. 
\texttt{MedCertAIn II} performs similarly but remains slightly lower across all four metrics. 
The corresponding selective prediction trends are shown in Appendix~\Cref{fig:figure_5}.



\subsection{Impact of Multimodal Fusion on Predictive and Selective Performance}

\setlength{\tabcolsep}{14.2pt}
\begin{table}[t!]
\vspace{-2mm}
\caption{\textbf{Impact of Multimodal Fusion on Predictive and Selective Performance.} Performance on 48-hour in-hospital mortality prediction using CXR-only, EHR-only, and multimodal inputs. Multimodal fusion yields the strongest overall performance, while \texttt{MedCertAIn} improves predictive and selective performance over MedFuse in the EHR-only and multimodal settings.}
\vspace{2mm}
\centering
\small
    \begin{tabular}{ll|cccc}
        \toprule
        \textbf{Modality} & \textbf{Model} & \textbf{AUROC} & \textbf{AUPRC} & \textbf{\makecell{Selective \\ AUROC}} & \textbf{\makecell{Selective \\ AUPRC}} \\
        \midrule
        
        \multirow{2}{*}{Unimodal CXR}
        & MedFuse & 0.615\tiny{$\pm$0.007} & 0.225\tiny{$\pm$0.007} & \cellcolor[gray]{0.85}\textbf{0.569\tiny{$\pm$0.012}} & \cellcolor[gray]{0.85}\textbf{0.167\tiny{$\pm$0.015}} \\
        & \texttt{MedCertAIn} & \cellcolor[gray]{0.85}\textbf{0.632\tiny{$\pm$0.014}} & \cellcolor[gray]{0.85}\textbf{0.242\tiny{$\pm$0.014}} & 0.567\tiny{$\pm$0.015} & 0.146\tiny{$\pm$0.004} \\
        \midrule
        
        \multirow{2}{*}{Unimodal EHR}
        & MedFuse & 0.760\tiny{$\pm$0.003} & 0.393\tiny{$\pm$0.008} & 0.670\tiny{$\pm$0.017} & 0.235\tiny{$\pm$0.007} \\
        & \texttt{MedCertAIn} & \cellcolor[gray]{0.85}\textbf{0.825\tiny{$\pm$0.004}} & \cellcolor[gray]{0.85}\textbf{0.480\tiny{$\pm$0.004}} & \cellcolor[gray]{0.85}\textbf{0.807\tiny{$\pm$0.014}} & \cellcolor[gray]{0.85}\textbf{0.499\tiny{$\pm$0.023}} \\
        \midrule
        
        \multirow{2}{*}{Multimodal}
        & MedFuse & 0.819\tiny{$\pm$0.000} & 0.466\tiny{$\pm$0.002} & 0.659\tiny{$\pm$0.005} & 0.203\tiny{$\pm$0.006} \\
        & \texttt{MedCertAIn} & \cellcolor[gray]{0.85}\textbf{0.835\tiny{$\pm$0.001}} & \cellcolor[gray]{0.85}\textbf{0.498\tiny{$\pm$0.002}} & \cellcolor[gray]{0.85}\textbf{0.857\tiny{$\pm$0.005}} & \cellcolor[gray]{0.85}\textbf{0.599\tiny{$\pm$0.002}} \\
        \bottomrule
    \end{tabular}
\label{table:table_5}
\end{table}

To quantify the benefit of fusion, \Cref{table:table_5} compares unimodal (CXR-only and EHR-only) and multimodal performance for the deterministic MedFuse baseline and our stochastic \texttt{MedCertAIn} framework. 
In its unimodal variants, \texttt{MedCertAIn} uses random-corruption context sets without latent-space similarity-based selection. 
Across both frameworks, EHR provides the strongest unimodal signal (MedFuse: AUROC 0.760, selective AUROC 0.670; \texttt{MedCertAIn}: AUROC 0.825, selective AUROC 0.807), while CXR-only performance remains lower (MedFuse: AUROC 0.615; \texttt{MedCertAIn}: AUROC 0.632).
Combining both modalities improves standard predictive performance: MedFuse achieves an AUROC of $0.819$ and an AUPRC of $0.466$, while \texttt{MedCertAIn} achieves an AUROC of $0.835$ and an AUPRC of $0.498$. 
For \texttt{MedCertAIn}, the largest gains appear in selective prediction, where the multimodal model achieves a selective AUROC of $0.857$ and a selective AUPRC of $0.599$, outperforming both the EHR-only ($0.807/0.499$) and CXR-only ($0.567/0.146$) variants.

\setlength{\tabcolsep}{5.4pt}
\begin{table*}[t!]
\caption{\textbf{Analysis of \texttt{MedCertAIn} Across Patient Subpopulations.} Comparison of MedFuse and \texttt{MedCertAIn} across age and sex subgroups. \texttt{MedCertAIn} improves AUROC and both selective metrics across all subgroups, while standard AUPRC varies across age groups.}

\vspace*{2mm}
    \resizebox{\textwidth}{!}{
    \small
    \begin{tabular}{l|cccc|cccc}
    \toprule
     \multirow{2}{*}[-0.5em]{\textbf{\makecell{Patient \\ Subgroup}}}
    & \multicolumn{4}{c}{\textbf{MedFuse}}
    & \multicolumn{4}{c}{\textbf{MedCertAIn}}\\

    & \textbf{AUROC}
    & \textbf{AUPRC} 
    & \textbf{\makecell{Selective \\ AUROC}}
    & \textbf{\makecell{Selective \\ AUPRC}}
    
    & \textbf{AUROC}
    & \textbf{AUPRC} 
    & \textbf{\makecell{Selective \\ AUROC}}
    & \textbf{\makecell{Selective \\ AUPRC}} \\
    
    \midrule
    \multicolumn{1}{c}{} & \multicolumn{8}{c}{\bfseries Age}\\
    \midrule

    18-45 years & 0.803\tiny{$\pm$0.001} & \cellcolor[gray]{0.85}\textbf{0.603\tiny{$\pm$0.023}} & 0.559\tiny{$\pm$0.002} & 0.262\tiny{$\pm$0.001} & \cellcolor[gray]{0.85}\textbf{0.824\tiny{$\pm$0.001}} & 0.600\tiny{$\pm$0.002} & \cellcolor[gray]{0.85}\textbf{0.723\tiny{$\pm$0.005}} & \cellcolor[gray]{0.85}\textbf{0.387\tiny{$\pm$0.017}}\\
    
    45-60 years & 0.788\tiny{$\pm$0.001} & \cellcolor[gray]{0.85}\textbf{0.545\tiny{$\pm$0.027}} & 0.554\tiny{$\pm$0.003} & 0.231\tiny{$\pm$0.001} & \cellcolor[gray]{0.85}\textbf{0.811\tiny{$\pm$0.002}} & 0.485\tiny{$\pm$0.002} & \cellcolor[gray]{0.85}\textbf{0.730\tiny{$\pm$0.008}} & \cellcolor[gray]{0.85}\textbf{0.299\tiny{$\pm$0.014}}\\
    
    $>$60 years & 0.814\tiny{$\pm$0.000} & \cellcolor[gray]{0.85}\textbf{0.528\tiny{$\pm$0.032}} & 0.631\tiny{$\pm$0.004} & 0.207\tiny{$\pm$0.002} & \cellcolor[gray]{0.85}\textbf{0.832\tiny{$\pm$0.000}} & 0.480\tiny{$\pm$0.002} & \cellcolor[gray]{0.85}\textbf{0.863\tiny{$\pm$0.004}} & \cellcolor[gray]{0.85}\textbf{0.559\tiny{$\pm$0.002}}\\

    \midrule
    \multicolumn{1}{c}{} & \multicolumn{8}{c}{\bfseries Sex}\\
    \midrule
    Male & 0.792\tiny{$\pm$0.000} & 0.401\tiny{$\pm$0.001} & 0.603\tiny{$\pm$0.003} & 0.186\tiny{$\pm$0.001} & \cellcolor[gray]{0.85}\textbf{0.810\tiny{$\pm$0.001}} & \cellcolor[gray]{0.85}\textbf{0.425\tiny{$\pm$0.002}} & \cellcolor[gray]{0.85}\textbf{0.832\tiny{$\pm$0.005}} & \cellcolor[gray]{0.85}\textbf{0.494\tiny{$\pm$0.005}}\\
    Female & 0.806\tiny{$\pm$0.000} & 0.420\tiny{$\pm$0.001} & 0.609\tiny{$\pm$0.003} & 0.183\tiny{$\pm$0.000} & \cellcolor[gray]{0.85}\textbf{0.827\tiny{$\pm$0.001}} & \cellcolor[gray]{0.85}\textbf{0.453\tiny{$\pm$0.002}} & \cellcolor[gray]{0.85}\textbf{0.842\tiny{$\pm$0.005}} & \cellcolor[gray]{0.85}\textbf{0.484\tiny{$\pm$0.005}}\\

    \bottomrule
    \end{tabular}
    }
    \label{table:table_3}
\end{table*}

\begin{figure*}[t]
    \vspace*{-2mm}
    \centering
    \includegraphics[width=\textwidth]{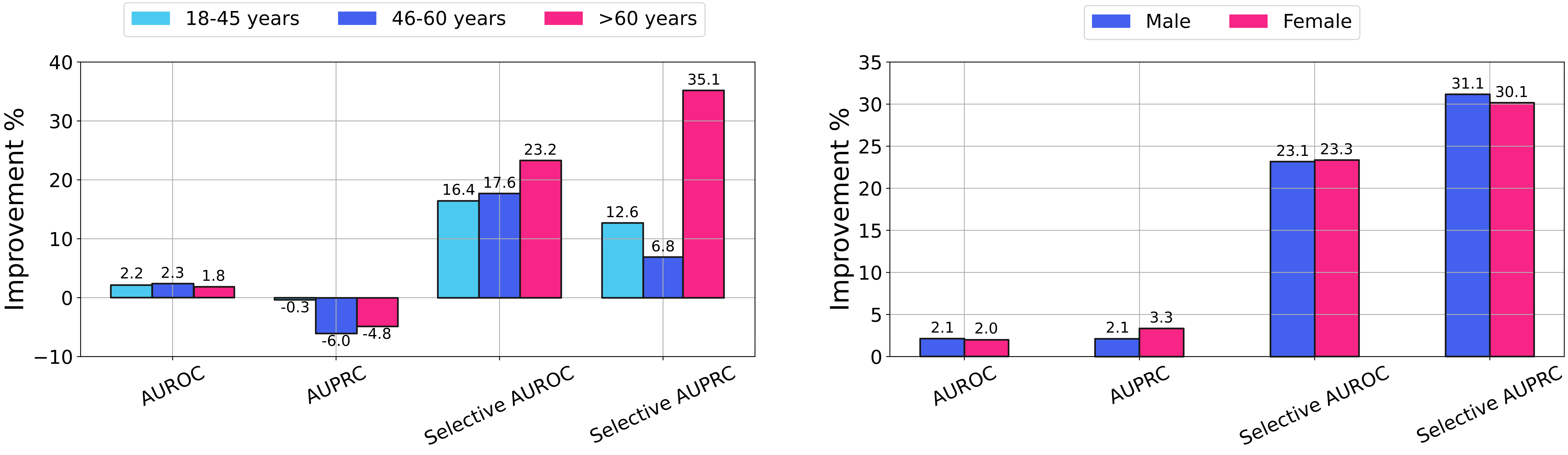}
    \vspace*{-20pt}
        \caption{\small
        \textbf{Relative Improvements Across Patient Subpopulations.}
        Percentage improvement of \texttt{MedCertAIn} over MedFuse across age and sex subgroups, highlighting the largest gains in selective prediction.}
        \vspace*{-8pt}
    \label{fig:figure_4}
\end{figure*}

\subsection{Analysis of \texttt{MedCertAIn} Across Patient Subpopulations}

To evaluate performance across patient subpopulations, we stratify the results by age and sex. 
\Cref{table:table_3} shows that \texttt{MedCertAIn} improves AUROC and both selective metrics over the deterministic MedFuse baseline across all subgroups. 
Across age groups, AUROC increases from 0.803 to 0.824 (18--45), 0.788 to 0.811 (45--60), and 0.814 to 0.832 ($>$60), while selective AUROC increases from 0.559 to 0.723, 0.554 to 0.730, and 0.631 to 0.863, respectively. 
Although standard AUPRC is slightly lower in the age subgroups, selective AUPRC consistently improves, increasing from 0.262 to 0.387 (18--45), 0.231 to 0.299 (45--60), and 0.207 to 0.559 ($>$60).

Across sex subgroups, \texttt{MedCertAIn} improves all four metrics for both male and female patients. 
For males, AUROC increases from 0.792 to 0.810 and selective AUROC from 0.603 to 0.832, with selective AUPRC improving from 0.186 to 0.494. 
For females, AUROC increases from 0.806 to 0.827 and selective AUROC from 0.609 to 0.842, with selective AUPRC improving from 0.183 to 0.484. 
The relative improvements across subgroups are visualized in \Cref{fig:figure_4}.
Overall, \texttt{MedCertAIn} maintains strong predictive performance while providing substantial gains in selective prediction across the evaluated subgroups, motivating further evaluation in diverse patient populations and clinically realistic deferral settings. 

\section{Discussion}
\label{sec:discussion}

Recent studies show that UQ methods in healthcare largely rely on Bayesian approaches but are often limited to unimodal imaging settings \citep{seoni2023application, uq_survey_lechuga}.
Our work addresses this gap by applying Bayesian UQ to multimodal clinical data combining medical images and time-series signals.
A key strength of \texttt{MedCertAIn} is that context-sample selection is performed without class labels, making it scalable across tasks and modalities.
Across experiments and subpopulation analyses, \texttt{MedCertAIn} achieves competitive predictive performance and consistently improves selective prediction, enabling more effective uncertainty-based ranking than the evaluated deterministic and stochastic baselines.

Despite these promising results, several limitations remain.
First, although we evaluate multiple time horizons, our experiments focus exclusively on mortality prediction, and generalization to other clinical tasks requires further study.
Second, incorporating additional modalities could improve applicability and performance \citep{salvi2023multi}.
The empirical gains from cross-modal similarity are modest in the current setting, although this signal provides a label-free and reusable proxy for multimodal disagreement.
Stronger task-adapted or clinically fine-tuned multimodal self-supervised models may improve the utility of similarity-based context selection.
Furthermore, improvements in selective prediction do not uniformly translate to better calibration across all metrics and prediction horizons.
Finally, evaluation using MIMIC-IV and MIMIC-CXR limits evidence of generalizability across institutions and patient populations, motivating external validation in more diverse clinical settings \citep{meng2022interpretability}.

Beyond performance gains, our findings suggest that uncertainty-aware multimodal models may offer practical benefits for clinical decision-making.
Improved selective prediction indicates the potential to defer uncertain cases, which is particularly relevant in high-risk ICU settings.
However, the rejection thresholds evaluated in this study are not fixed clinical decision thresholds; a deployment-specific operating point would depend on acceptable risk and available review capacity.
Moreover, Shannon entropy reflects uncertainty in the learned Bayesian predictive distribution and therefore depends on the model architecture, variational posterior, and prior.
Consistent selective improvements across subpopulations are encouraging, but do not establish fairness and warrant dedicated investigation across broader demographic and clinical groups.

Building on these results, future work should explore additional modalities such as radiology reports, extend evaluation to other clinical tasks such as decompensation and length-of-stay prediction, and investigate alternative fusion strategies, including late fusion and missing-modality settings.
Strengthening the multimodal backbone and exploring improved self-supervised representations for identifying high-uncertainty samples may further improve uncertainty estimation.

\section{Conclusion}
\label{sec:conclusion}

We introduced \texttt{MedCertAIn}, a Bayesian uncertainty quantification framework that uses data-driven priors for multimodal in-hospital mortality prediction from chest X-ray images and EHR time-series data.
Across experiments, \texttt{MedCertAIn} achieves competitive predictive performance and substantial gains in selective prediction over the evaluated deterministic and stochastic baselines.
These results highlight the potential of data-driven Bayesian multimodal learning to improve uncertainty-based ranking and support the deferral of uncertain cases for clinical review.



\subsubsection*{Acknowledgments}

This work was supported by the NYUAD Center for
Artificial Intelligence and Robotics, funded by Tamkeen under the NYUAD Research Institute Award
CG010. 
The research was carried out on the High
Performance Computing resources at New York University Abu Dhabi.

\bibliography{references}
\bibliographystyle{tmlr}

\clearpage

\appendix

\crefalias{section}{appendix}
\crefalias{subsection}{appendix}
\crefname{appendix}{Appendix}{Appendices}
\Crefname{appendix}{Appendix}{Appendices}

\renewcommand{\thesection}{\Alph{section}}
\renewcommand{\thesubsection}{\thesection.\arabic{subsection}}

\section*{\LARGE Appendix}
\label{sec:appendix}

\setcounter{table}{0}
\setcounter{figure}{0}
\setcounter{equation}{0}
\setcounter{section}{0}
\setcounter{subsection}{0}
\renewcommand{\thetable}{\thesection.\arabic{table}}
\renewcommand{\thefigure}{\thesection.\arabic{figure}}
\renewcommand{\theequation}{\thesection.\arabic{equation}}
\renewcommand{\theHfigure}{\thesection.\arabic{figure}}

\vspace*{10pt}
\section*{Table of Contents}
\startcontents[sections]
\printcontents[sections]{l}{1}{\setcounter{tocdepth}{2}}

\clearpage
\section{Extended Results}
\label{sec:extended_results}

\subsection{48-Hour In-Hospital Mortality}
\setlength{\tabcolsep}{10pt}
\begin{table}[h]
\caption{\textbf{48-Hour Mortality Calibration.} Calibration across MedFuse-based baselines and \texttt{MedCertAIn}. \texttt{MedCertAIn} achieves the lowest Brier score, whereas MedFuse obtains the lowest ECE.}
\vspace{2mm}
    \small
    \centering
    \begin{tabular}{lcc}
    
    \toprule
    
    \diagbox[width=9em]{\textbf{Model}}{\textbf{Metric}} 
    & \textbf{ECE $\downarrow$}
    & \textbf{Brier $\downarrow$} \\
    
    \midrule
    \multicolumn{3}{c}{\textbf{MedFuse Variations}} \\
    \midrule
   
   MedFuse \citep{medfuse} 
   & \cellcolor[gray]{0.85}\textbf{3.215\tiny{$\pm$0.242}}
   & \underline{0.102\tiny{$\pm$0.000}} \\[3pt]

   Temp. Scaling \citep{guo2017calibration} 
   & 3.527\tiny{$\pm$1.225}
   & 0.102\tiny{$\pm$0.001} \\[3pt]

   Deep Ensemble \citep{lakshminarayanan2016simple} 
   & \underline{3.455\tiny{$\pm$0.000}}
   & 0.102\tiny{$\pm$0.000} \\[3pt]

   MC Dropout \citep{gal2016dropoutbayesianapproximationrepresenting} 
   & 3.613\tiny{$\pm$0.424}
   & 0.102\tiny{$\pm$0.001} \\[3pt]

   Uninformative Prior \citep{fsvi} 
   & 4.625\tiny{$\pm$0.929}
   & 0.102\tiny{$\pm$0.001} \\[3pt]

   \midrule
   \textbf{\texttt{MedCertAIn}} (\textbf{Ours}) 
   & 4.323\tiny{$\pm$0.146}
   & \cellcolor[gray]{0.85}\textbf{0.100\tiny{$\pm$0.000}} \\

    \bottomrule
    \end{tabular}
   \label{table:48_hour_calibration}
    
\end{table}

\Cref{table:48_hour_calibration} reports calibration performance for 48-hour mortality prediction. 
We note that in this setting, \texttt{MedCertAIn} achieves the lowest Brier score, indicating the strongest overall probabilistic accuracy among the evaluated methods. However, MedFuse obtains the lowest ECE, while \texttt{MedCertAIn} has the second-highest ECE. 
These results reveal a trade-off: \texttt{MedCertAIn} improves the Brier score but does not improve calibration as measured by ECE.
The prediction performance results for this prediction task are reported in \Cref{table:main_table} in the main text.

\clearpage
\subsection{3-Month In-Hospital Mortality}
\setlength{\tabcolsep}{8pt}
\begin{table}[h]
\caption{\textbf{3-Month Mortality Prediction Performance.} Comparison across MedFuse-based baselines and \texttt{MedCertAIn}, which achieves the highest mean performance across all standard and selective metrics, with the largest improvement in selective AUPRC.}

\vspace{2mm}
    \small
    \centering
    \begin{tabular}{lcccc}
    
    \toprule
    
    \diagbox[width=9em]{\textbf{Model}}{\textbf{Metric}} 
    & \textbf{AUROC} 
    & \textbf{AUPRC} 
    & \textbf{\makecell{Selective \\ AUROC}} 
    & \textbf{\makecell{ Selective \\ AUPRC}}\\
    
    \midrule
    \multicolumn{5}{c}{\textbf{MedFuse Variations}} \\
    \midrule

   MedFuse \citep{medfuse} 
   & 0.656\tiny{$\pm$0.004} 
   & 0.174\tiny{$\pm$0.006}
   & 0.582\tiny{$\pm$0.009} 
   & 0.078\tiny{$\pm$0.002} \\[3pt]

   Temp. Scaling \citep{guo2017calibration} 
   & 0.656\tiny{$\pm$0.004} 
   & 0.174\tiny{$\pm$0.006}
   & 0.582\tiny{$\pm$0.009} 
   & 0.078\tiny{$\pm$0.002} \\[3pt]

   Deep Ensemble \citep{lakshminarayanan2016simple} 
   & \underline{0.657\tiny{$\pm$0.000}} 
   & \underline{0.180\tiny{$\pm$0.000}} 
   & \underline{0.583\tiny{$\pm$0.000}} 
   & 0.077\tiny{$\pm$0.000} \\[3pt]
    
   MC Dropout \citep{gal2016dropoutbayesianapproximationrepresenting} 
   & 0.656\tiny{$\pm$0.004} 
   & 0.175\tiny{$\pm$0.010} 
   & 0.556\tiny{$\pm$0.013} 
   & \underline{0.102\tiny{$\pm$0.002}} \\[3pt]

   Uninformative Prior \citep{fsvi} 
   & 0.651\tiny{$\pm$0.021} 
   & 0.146\tiny{$\pm$0.021} 
   & 0.581\tiny{$\pm$0.031}
   & 0.079\tiny{$\pm$0.003} \\[3pt]

    \midrule
   \textbf{\texttt{MedCertAIn}} (\textbf{Ours}) 
   & \cellcolor[gray]{0.85}\textbf{0.660\tiny{$\pm$0.001}} 
   & \cellcolor[gray]{0.85}\textbf{0.183\tiny{$\pm$0.009}} 
   & \cellcolor[gray]{0.85}\textbf{0.683\tiny{$\pm$0.020}} 
   & \cellcolor[gray]{0.85}\textbf{0.196\tiny{$\pm$0.018}} \\

       Improvement vs MedFuse (\%) 
    & \cellcolor{lightgreen}+0.6\% 
    & \cellcolor{lightgreen}+5.2\% 
    & \cellcolor{lightgreen}+17.4\% 
    & \cellcolor{lightgreen}+151.3\% \\
    
    Improvement vs Uninformative Prior (\%) 
    & \cellcolor{lightgreen}+1.4\% 
    & \cellcolor{lightgreen}+25.3\% 
    & \cellcolor{lightgreen}+17.6\% 
    & \cellcolor{lightgreen}+148.1\% \\
 
    \bottomrule
    \end{tabular}
   \label{table:3_month_performance}
    
\end{table}
\setlength{\tabcolsep}{10pt}
\begin{table}[h]
\caption{\textbf{3-Month Mortality Calibration.} Calibration and probabilistic accuracy across MedFuse-based baselines and \texttt{MedCertAIn}. Deep Ensemble achieves the lowest ECE, while MedFuse, Deep Ensemble, and MC Dropout share the lowest reported Brier score.}

\vspace{2mm}
    \small
    \centering
    \begin{tabular}{lcc}
    
    \toprule
    
    \diagbox[width=9em]{\textbf{Model}}{\textbf{Metric}} 
    & \textbf{ECE $\downarrow$}
    & \textbf{Brier $\downarrow$} \\

    \midrule
    \multicolumn{3}{c}{\textbf{MedFuse Variations}} \\
    \midrule
    
   MedFuse \citep{medfuse} 
   & 0.979\tiny{$\pm$0.347}
   & \cellcolor[gray]{0.85}\textbf{0.068\tiny{$\pm$0.000}} \\[3pt]

   Temp. Scaling \citep{guo2017calibration} 
   & 1.545\tiny{$\pm$0.137}
   & \underline{0.069\tiny{$\pm$0.000}} \\[3pt]

   Deep Ensemble \citep{lakshminarayanan2016simple} 
   & \cellcolor[gray]{0.85}\textbf{0.847\tiny{$\pm$0.000}}
   & \cellcolor[gray]{0.85}\textbf{0.068\tiny{$\pm$0.000}} \\[3pt]

   MC Dropout \citep{gal2016dropoutbayesianapproximationrepresenting}
   & \underline
   {0.937\tiny{$\pm$0.347}}
   & \cellcolor[gray]{0.85}\textbf{0.068\tiny{$\pm$0.000}} \\[3pt]

   Uninformative Prior \citep{fsvi}
   & 1.313\tiny{$\pm$1.036}
   & 0.070\tiny{$\pm$0.004} \\[3pt]

   \midrule
   \textbf{\texttt{MedCertAIn}} (\textbf{Ours})
   & 1.492\tiny{$\pm$0.049}
   & 0.076\tiny{$\pm$0.001} \\

    \bottomrule
    \end{tabular}
   \label{table:mortality_3month_calibration}
    
\end{table}

\Cref{table:3_month_performance,table:mortality_3month_calibration} report predictive, selective, and calibration performance for 3-month mortality prediction.
For 3-month mortality prediction, \texttt{MedCertAIn} achieves the highest mean AUROC, AUPRC, selective AUROC, and selective AUPRC among the evaluated methods. 
The largest gain occurs in selective AUPRC, which improves by $151.3\%$ over MedFuse, indicating more effective uncertainty-based ranking. 
Calibration results show a different pattern: Deep Ensemble obtains the lowest ECE, while MedFuse, Deep Ensemble, and MC Dropout share the lowest reported Brier score. 
Thus, \texttt{MedCertAIn}'s substantial selective prediction gains do not translate to improved calibration at this prediction horizon.

\clearpage
\subsection{6-Month In-Hospital Mortality}
\setlength{\tabcolsep}{8pt}
\begin{table}[h]
\caption{\textbf{6-Month Mortality Prediction Performance.} Comparison across MedFuse-based baselines and \texttt{MedCertAIn}. \texttt{MedCertAIn} achieves the highest mean AUROC and substantially improves both selective metrics, although its AUPRC is lower than those of the evaluated baselines.}

\vspace{2mm}
    \small
    \centering
    \begin{tabular}{lcccc}
    
    \toprule
    
    \diagbox[width=9em]{\textbf{Model}}{\textbf{Metric}} 
    & \textbf{AUROC} 
    & \textbf{AUPRC} 
    & \textbf{\makecell{Selective \\ AUROC}} 
    & \textbf{\makecell{ Selective \\ AUPRC}}\\
    
    \midrule
    \multicolumn{5}{c}{\textbf{MedFuse Variations}} \\
    \midrule
   


   MedFuse \citep{medfuse} 
   & 0.642\tiny{$\pm$0.005} 
   & 0.196\tiny{$\pm$0.013} 
   & 0.544\tiny{$\pm$0.011} 
   & 0.102\tiny{$\pm$0.002} \\[3pt]

   Temp. Scaling \citep{guo2017calibration} 
   & 0.642\tiny{$\pm$0.005} 
   & 0.196\tiny{$\pm$0.013} 
   & 0.544\tiny{$\pm$0.011} 
   & 0.102\tiny{$\pm$0.002} \\[3pt]

   Deep Ensemble \citep{lakshminarayanan2016simple} 
   & 0.646\tiny{$\pm$0.000} 
   & \cellcolor[gray]{0.85}\textbf{0.211\tiny{$\pm$0.000}} 
   & 0.544\tiny{$\pm$0.000} 
   & 0.100\tiny{$\pm$0.000} \\[3pt]
    
   MC Dropout \citep{gal2016dropoutbayesianapproximationrepresenting} 
   & \underline{0.648\tiny{$\pm$0.002}}
   & \underline{0.205\tiny{$\pm$0.006}} 
   & 0.556\tiny{$\pm$0.013} 
   & 0.102\tiny{$\pm$0.002} \\[3pt]

   Uninformative Prior \citep{fsvi} 
   & 0.644\tiny{$\pm$0.003} 
   & 0.197\tiny{$\pm$0.009} 
   & \underline{0.563\tiny{$\pm$0.013}} 
   & \underline{0.103\tiny{$\pm$0.003}} \\[3pt]

   \midrule
   \textbf{\texttt{MedCertAIn}} (\textbf{Ours}) 
   & \cellcolor[gray]{0.85}\textbf{0.654\tiny{$\pm$0.005}} 
   & 0.189\tiny{$\pm$0.006} 
   & \cellcolor[gray]{0.85}\textbf{0.683\tiny{$\pm$0.020}} 
   & \cellcolor[gray]{0.85}\textbf{0.196\tiny{$\pm$0.027}} \\

   Improvement vs MedFuse (\%) 
   & \cellcolor{lightgreen}+1.9\% 
   & \cellcolor{lightred}-3.6\% 
   & \cellcolor{lightgreen}+25.6\% 
   & \cellcolor{lightgreen}+92.2\% \\

   Improvement vs Uninformative Prior (\%) 
   & \cellcolor{lightgreen}+1.6\% 
   & \cellcolor{lightred}-4.1\% 
   & \cellcolor{lightgreen}+21.3\% 
   & \cellcolor{lightgreen}+90.3\% \\

    \bottomrule
    \end{tabular}
   \label{table:6_month_performance}
    
\end{table}
\setlength{\tabcolsep}{10pt}
\begin{table}[h]
\caption{\textbf{6-Month Mortality Calibration.} Calibration and probabilistic accuracy across MedFuse-based baselines and \texttt{MedCertAIn}. Deep Ensemble achieves the lowest ECE, while Deep Ensemble, MC Dropout, and the uninformative-prior BNN share the lowest reported Brier score.}

\vspace{2mm}
    \small
    \centering
    \begin{tabular}{lcc}
    
    \toprule
    
    \diagbox[width=9em]{\textbf{Model}}{\textbf{Metric}} 
    & \textbf{ECE $\downarrow$}
    & \textbf{Brier $\downarrow$} \\
    
    \midrule
    \multicolumn{3}{c}{\textbf{MedFuse Variations}} \\
    \midrule
    
   MedFuse \citep{medfuse} 
   & 1.168\tiny{$\pm$0.451}
   & \underline{0.092\tiny{$\pm$0.000}} \\[3pt]

   Temp. Scaling \citep{guo2017calibration} 
   & \underline{0.831\tiny{$\pm$0.808}}
   & \underline{0.092\tiny{$\pm$0.000}} \\[3pt]

   Deep Ensemble \citep{lakshminarayanan2016simple} 
   & \cellcolor[gray]{0.85}\textbf{0.699\tiny{$\pm$0.000}}
   & \cellcolor[gray]{0.85}\textbf{0.091\tiny{$\pm$0.000}}  \\[3pt]

   MC Dropout \citep{gal2016dropoutbayesianapproximationrepresenting}
   & 1.452\tiny{$\pm$0.419}
   & \cellcolor[gray]{0.85}\textbf{0.091\tiny{$\pm$0.000}} \\[3pt]

   Uninformative Prior \citep{fsvi}
   & 2.065\tiny{$\pm$0.149}
   & \cellcolor[gray]{0.85}\textbf{0.091\tiny{$\pm$0.000}} \\[3pt]

   \midrule
   \textbf{\texttt{MedCertAIn}} (\textbf{Ours})
   & 2.105\tiny{$\pm$0.108}
   & \underline{0.092\tiny{$\pm$0.000}} \\

    \bottomrule
    \end{tabular}
   \label{table:mortality_6month_calibration}
    
\end{table}

\Cref{table:6_month_performance,table:mortality_6month_calibration} report predictive, selective, and calibration performance for 6-month mortality prediction.
In this setting, \texttt{MedCertAIn} achieves the highest mean AUROC and substantially improves both selective metrics relative to MedFuse and the uninformative-prior BNN. Compared with MedFuse, selective AUROC increases by $25.6\%$ and selective AUPRC by $92.2\%$, indicating more effective uncertainty-based ranking. 
However, \texttt{MedCertAIn} obtains a lower AUPRC and does not improve calibration, scoring the highest ECE among the evaluated methods. 
These results show that its primary benefit at this horizon lies in selective prediction, while standard precision--recall performance and calibration remain mixed.

\clearpage
\subsection{1-Year In-Hospital Mortality}
\setlength{\tabcolsep}{8pt}
\begin{table}[h]
\caption{\textbf{1-Year Mortality Prediction Performance.} Comparison across MedFuse-based baselines and \texttt{MedCertAIn}. \texttt{MedCertAIn} achieves the highest mean AUROC and selective prediction performance, with the largest improvement in selective AUPRC.}

\vspace{2mm}
    \small
    \centering
    \begin{tabular}{lcccc}
    
    \toprule
    
    \diagbox[width=9em]{\textbf{Model}}{\textbf{Metric}} & \textbf{AUROC} & \textbf{AUPRC} & \textbf{\makecell{Selective \\ AUROC}} & \textbf{\makecell{ Selective \\ AUPRC}}\\
    
    \midrule
    \multicolumn{5}{c}{\textbf{MedFuse Variations}} \\
    \midrule
   
   MedFuse \citep{medfuse} & 0.636\tiny{$\pm$0.005} & 0.227\tiny{$\pm$0.007} & \underline{0.545\tiny{$\pm$0.009}} & 0.137\tiny{$\pm$0.002} \\[3pt]

   Temp. Scaling \citep{guo2017calibration} & 0.636\tiny{$\pm$0.005} & 0.227\tiny{$\pm$0.007} & \underline{0.545\tiny{$\pm$0.009}} & 0.137\tiny{$\pm$0.002} \\[3pt]

   Deep Ensemble
   \citep{lakshminarayanan2016simple} & \underline{0.638\tiny{$\pm$0.000}} & \underline{0.230\tiny{$\pm$0.000}} & 0.544\tiny{$\pm$0.000} & 0.136\tiny{$\pm$0.000} \\[3pt]

   MC Dropout \citep{gal2016dropoutbayesianapproximationrepresenting} & 0.637\tiny{$\pm$0.003} & \cellcolor[gray]{0.85}\textbf{0.231\tiny{$\pm$0.005}} & \underline{0.545\tiny{$\pm$0.014}} & 0.135\tiny{$\pm$0.001} \\[3pt]

   Uninformative Prior \citep{fsvi} & 0.593\tiny{$\pm$0.023} & 0.197\tiny{$\pm$0.014} & 0.537\tiny{$\pm$0.031} & \underline{0.167\tiny{$\pm$0.020}} \\[3pt]
    
   \midrule
   \textbf{\texttt{MedCertAIn}} (\textbf{Ours}) & \cellcolor[gray]{0.85}\textbf{0.639\tiny{$\pm$0.004}} & 0.222\tiny{$\pm$0.005} & \cellcolor[gray]{0.85}\textbf{0.661\tiny{$\pm$0.011}} & \cellcolor[gray]{0.85}\textbf{0.252\tiny{$\pm$0.015}} \\

    Improvement vs MedFuse (\%) & \cellcolor{lightgreen}+0.5\% & \cellcolor{lightred}-2.2\% & \cellcolor{lightgreen}+21.3\% & \cellcolor{lightgreen}+83.9\% \\

   Improvement vs Uninformative Prior (\%) & \cellcolor{lightgreen}+7.8\% & \cellcolor{lightgreen}+12.7\% & \cellcolor{lightgreen}+23.1\% & \cellcolor{lightgreen}+50.9\% \\
   
    \bottomrule
    \end{tabular}
   \label{table:1_year_performance}
    
\end{table}
\setlength{\tabcolsep}{10pt}
\begin{table}[h]
\caption{\textbf{1-Year Mortality Calibration.} Calibration and probabilistic accuracy across MedFuse-based baselines and \texttt{MedCertAIn}. MC Dropout achieves the lowest ECE, while all MedFuse-based baselines except the uninformative-prior BNN share the lowest reported Brier score.}

\vspace{2mm}
    \small
    \centering
    \begin{tabular}{lcc}
    
    \toprule
    
    \diagbox[width=9em]{\textbf{Model}}{\textbf{Metric}} 
    & \textbf{ECE $\downarrow$}
    & \textbf{Brier $\downarrow$} \\
   
    \midrule
    \multicolumn{3}{c}{\textbf{MedFuse Variations}} \\
    \midrule
    
   MedFuse \citep{medfuse} 
   & 2.086\tiny{$\pm$0.430}
   & \cellcolor[gray]{0.85}\textbf{0.118\tiny{$\pm$0.000}} \\[3pt]

   Temp. Scaling \citep{guo2017calibration} 
   & \underline{2.007\tiny{$\pm$0.502}}
   & \cellcolor[gray]{0.85}\textbf{0.118\tiny{$\pm$0.000}} \\[3pt]

   Deep Ensemble \citep{lakshminarayanan2016simple} 
   & 2.718\tiny{$\pm$0.000}
   & \cellcolor[gray]{0.85}\textbf{0.118\tiny{$\pm$0.000}} \\[3pt]

   MC Dropout \citep{gal2016dropoutbayesianapproximationrepresenting}
   & \cellcolor[gray]{0.85}\textbf{1.879\tiny{$\pm$0.622}}
   & \cellcolor[gray]{0.85}\textbf{0.118\tiny{$\pm$0.000}} \\[3pt]

   Uninformative Prior \citep{fsvi}
   & 5.447\tiny{$\pm$2.330}
   & 0.135\tiny{$\pm$0.011} \\[3pt]

   \midrule
   \textbf{\texttt{MedCertAIn}} (\textbf{Ours})
   & 2.399\tiny{$\pm$0.120}
   & \underline{0.119\tiny{$\pm$0.001}} \\

    \bottomrule
    \end{tabular}
   \label{table:mortality_1year_calibration}
    
\end{table}

\Cref{table:1_year_performance,table:mortality_1year_calibration} report predictive, selective, and calibration performance for 1-year mortality prediction.
In this longer time horizon setting, \texttt{MedCertAIn} achieves the highest mean AUROC and substantially improves both selective metrics. Relative to MedFuse, selective AUROC increases by $21.3\%$ and selective AUPRC by $83.9\%$, indicating more effective uncertainty-based ranking. However, its AUPRC is lower than those of several MedFuse-based baselines. 
Calibration results show a similar trade-off: MC Dropout obtains the lowest ECE, while \texttt{MedCertAIn} achieves a slightly higher Brier score than the MedFuse-based baselines. Overall, the primary gains at this horizon occur in selective prediction rather than precision--recall performance or calibration.

\clearpage
\subsection{Comparison Across Deterministic Multimodal Models}
\label{sec:deterministic_models_comparison}
\setlength{\tabcolsep}{12pt}
\begin{table}[h]

\caption{\textbf{Deterministic Multimodal Models Across Mortality Horizons.} Performance of DrFuse, MeTra, and MedFuse on 48-hour, 3-month, 6-month, and 1-year mortality prediction. We report standard, selective, and calibration metrics as mean $\pm$ standard error across five random seeds. \footnotesize{The best and second-best results for each horizon and metric are shown in bold and underlined, respectively.}}

\vspace{2mm}
\small
\centering
\renewcommand{\arraystretch}{1.15}
\resizebox{\linewidth}{!}{
\begin{tabular}{l|cccc|cc}

\toprule

\textbf{Model}
& \textbf{AUROC}
& \textbf{AUPRC}
& \textbf{\makecell{Selective \\ AUROC}}
& \textbf{\makecell{Selective \\ AUPRC}}
& \textbf{ECE $\downarrow$}
& \textbf{Brier $\downarrow$} \\

\midrule
\multicolumn{7}{c}{\textbf{48h Mortality}} \\
\midrule

DrFuse 
& \underline{0.817\tiny{$\pm$0.005}}
& \cellcolor[gray]{0.85}\textbf{0.472\tiny{$\pm$0.013}}
& \cellcolor[gray]{0.85}\textbf{0.785\tiny{$\pm$0.007}}
& \cellcolor[gray]{0.85}\textbf{0.437\tiny{$\pm$0.035}}
& 8.943\tiny{$\pm$0.597}
& 0.111\tiny{$\pm$0.006} \\[3pt]

MeTra 
& 0.783\tiny{$\pm$0.015}
& 0.411\tiny{$\pm$0.037}
& \underline{0.724\tiny{$\pm$0.034}}
& \underline{0.296\tiny{$\pm$0.083}}
& \cellcolor[gray]{0.85}\textbf{2.062\tiny{$\pm$0.918}}
& \underline{0.110\tiny{$\pm$0.005}} \\[3pt]

MedFuse
& \cellcolor[gray]{0.85}\textbf{0.819\tiny{$\pm$0.000}}
& \underline{0.466\tiny{$\pm$0.002}}
& 0.659\tiny{$\pm$0.005}
& 0.203\tiny{$\pm$0.006}
& \underline{3.215\tiny{$\pm$0.242}}
& \cellcolor[gray]{0.85}\textbf{0.102\tiny{$\pm$0.000}} \\[3pt]

\midrule
\multicolumn{7}{c}{\textbf{3-month Mortality}} \\
\midrule

DrFuse 
& \underline{0.647\tiny{$\pm$0.039}}
& \underline{0.150\tiny{$\pm$0.035}}
& 0.574\tiny{$\pm$0.034}
& \cellcolor[gray]{0.85}\textbf{0.091\tiny{$\pm$0.024}}
& 4.159\tiny{$\pm$2.585}
& 0.079\tiny{$\pm$0.018} \\[3pt]

MeTra 
& 0.596\tiny{$\pm$0.028}
& 0.109\tiny{$\pm$0.015}
& \cellcolor[gray]{0.85}\textbf{0.588\tiny{$\pm$0.038}}
& \underline{0.085\tiny{$\pm$0.008}}
& \underline{1.756\tiny{$\pm$0.924}}
& \underline{0.071\tiny{$\pm$0.002}} \\[3pt]

MedFuse
& \cellcolor[gray]{0.85}\textbf{0.656\tiny{$\pm$0.004}}
& \cellcolor[gray]{0.85}\textbf{0.174\tiny{$\pm$0.006}}
& \underline{0.582\tiny{$\pm$0.009}}
& 0.078\tiny{$\pm$0.002}
& \cellcolor[gray]{0.85}\textbf{0.979\tiny{$\pm$0.347}}
& \cellcolor[gray]{0.85}\textbf{0.068\tiny{$\pm$0.000}} \\[3pt]

\midrule
\multicolumn{7}{c}{\textbf{6-month Mortality}} \\
\midrule

DrFuse 
& \cellcolor[gray]{0.85}\textbf{0.687\tiny{$\pm$0.011}}
& \cellcolor[gray]{0.85}\textbf{0.230\tiny{$\pm$0.008}}
& \underline{0.551\tiny{$\pm$0.018}}
& 0.098\tiny{$\pm$0.003}
& \underline{2.677\tiny{$\pm$1.512}}
& \cellcolor[gray]{0.85}\textbf{0.090\tiny{$\pm$0.002}} \\[3pt]

MeTra 
& 0.594\tiny{$\pm$0.030}
& 0.146\tiny{$\pm$0.015}
& \cellcolor[gray]{0.85}\textbf{0.562\tiny{$\pm$0.044}}
& \cellcolor[gray]{0.85}\textbf{0.112\tiny{$\pm$0.007}}
& 2.966\tiny{$\pm$1.530}
& 0.095\tiny{$\pm$0.003} \\[3pt]

MedFuse
& \underline{0.642\tiny{$\pm$0.005}}
& \underline{0.196\tiny{$\pm$0.013}}
& 0.544\tiny{$\pm$0.011}
& \underline{0.102\tiny{$\pm$0.002}}
& \cellcolor[gray]{0.85}\textbf{1.168\tiny{$\pm$0.451}}
& \underline{0.092\tiny{$\pm$0.000}} \\[3pt]

\midrule
\multicolumn{7}{c}{\textbf{1-year Mortality}} \\
\midrule

DrFuse 
& \cellcolor[gray]{0.85}\textbf{0.663\tiny{$\pm$0.009}}
& \cellcolor[gray]{0.85}\textbf{0.257\tiny{$\pm$0.005}}
& \underline{0.572\tiny{$\pm$0.015}}
& 0.134\tiny{$\pm$0.003}
& 3.501\tiny{$\pm$2.808}
& 0.119\tiny{$\pm$0.005} \\[3pt]

MeTra 
& 0.626\tiny{$\pm$0.002}
& 0.202\tiny{$\pm$0.010}
& \cellcolor[gray]{0.85}\textbf{0.595\tiny{$\pm$0.021}}
& \cellcolor[gray]{0.85}\textbf{0.144\tiny{$\pm$0.005}}
& \cellcolor[gray]{0.85}\textbf{1.906\tiny{$\pm$1.514}}
& \underline{0.119\tiny{$\pm$0.001}} \\[3pt]

MedFuse
& \underline{0.636\tiny{$\pm$0.005}}
& \underline{0.227\tiny{$\pm$0.007}}
& 0.545\tiny{$\pm$0.009}
& \underline{0.137\tiny{$\pm$0.002}}
& \underline{2.086\tiny{$\pm$0.430}}
& \cellcolor[gray]{0.85}\textbf{0.118\tiny{$\pm$0.000}} \\[3pt]

\bottomrule
\end{tabular}
}
\label{table:deterministic_models_merged}

\end{table}

\Cref{table:deterministic_models_merged} compares the deterministic multimodal architectures across all mortality horizons.
We compare three deterministic multimodal architectures with distinct fusion strategies. 
MedFuse \citep{medfuse} uses modality-specific encoders followed by concatenation-based early fusion. 
DrFuse \citep{yao2024drfuse} disentangles shared and modality-specific representations and dynamically weights the contribution of each modality. 
MeTra \citep{khader2023medical} represents imaging and clinical inputs as tokens and uses transformer attention for multimodal integration.

Across these baselines, no single model dominates all mortality horizons and evaluation criteria. MedFuse performs strongly at shorter horizons, achieving the highest AUROC at 48 hours and the highest AUROC and AUPRC at 3 months. DrFuse achieves the strongest standard predictive performance at 6 months and 1 year. 
Selective performance follows a different pattern: DrFuse performs best at 48 hours, while performance is split between DrFuse and MeTra at 3 months and favors MeTra at 6 months and 1 year. 
Calibration also varies by horizon, with MedFuse or MeTra generally achieving the lowest ECE and MedFuse or DrFuse the lowest Brier score. 
Overall, these results demonstrate trade-offs among discrimination, uncertainty-based ranking, and calibration, with no deterministic architecture performing consistently best across all criteria.

\clearpage

\subsection{Comparison with Hard Example Mining}
\setlength{\tabcolsep}{5pt}
\begin{table}[h]
    \caption{\small \textbf{Comparison with Hard Example Mining.} Performance when high-loss samples are used for direct HEM fine-tuning of MedFuse or as context for the uncertainty-aware prior in \texttt{MedCertAIn}.}

    \label{table:table_4}
    \small
    \vspace*{5pt}
    \centering
    \resizebox{\linewidth}{!}{
    \begin{tabular}{lcccc}
    \toprule
    \textbf{Model} & \textbf{AUROC} & \textbf{AUPRC} & \textbf{\makecell{Selective \\ AUROC}} & \textbf{\makecell{Selective \\ AUPRC}} \\
    \midrule
    MedFuse HEM \citep{shrivastava2016training}   & 
    0.233\tiny{$\pm$0.003} & 0.094\tiny{$\pm$0.001} & 0.393\tiny{$\pm$0.010} & 0.070\tiny{$\pm$0.010} \\
    
    MedCertAIn (HEM 20\%)   & \cellcolor[gray]{0.85}\textbf{0.821\tiny{$\pm$0.002}} & \cellcolor[gray]{0.85}\textbf{0.477\tiny{$\pm$0.002}} & \cellcolor[gray]{0.85}\textbf{0.790\tiny{$\pm$0.011}} & \cellcolor[gray]{0.85}\textbf{0.502\tiny{$\pm$0.023}} \\
    
    Absolute Difference (pp) & +58.75 & +38.30 & +39.65 & +43.26 \\
    
    \bottomrule
    \end{tabular}
    }
\end{table}

\Cref{table:table_4} reports the results of an additional comparison between our approach and Hard Example Mining (HEM), which prioritizes high-loss examples during training \citep{shrivastava2016training}. 
We adapt HEM fine-tuning using high-loss samples as context for the uncertainty-aware prior.
Direct HEM fine-tuning substantially degrades MedFuse performance. 
In a separate ablation, we instead use the top 20\% highest-loss training samples as the context set for constructing \texttt{MedCertAIn}'s uncertainty-aware prior.

Using hard examples as prior context achieves an AUROC of $0.821$, an AUPRC of $0.477$, a selective AUROC of $0.790$, and a selective AUPRC of $0.502$. 
Compared with direct HEM fine-tuning, these results correspond to absolute improvements of $58.75$, $38.30$, $39.65$, and $43.26$ percentage points, respectively. 
Nevertheless, this variant remains below the primary \texttt{MedCertAIn} configuration reported in \Cref{table:main_table}.

These results illustrate the distinction between the two approaches. 
HEM trains the predictive model directly on difficult samples, whereas this \texttt{MedCertAIn} variant uses them to shape the uncertainty-aware prior. 
The training objective continues to fit the model using the full training set while encouraging uncertainty on the hard-example context set.

\clearpage
\subsection{Comparison Across Multimodal Data-Driven Priors}
\begin{figure}[h!]
    \centering
    \includegraphics[width=\linewidth]{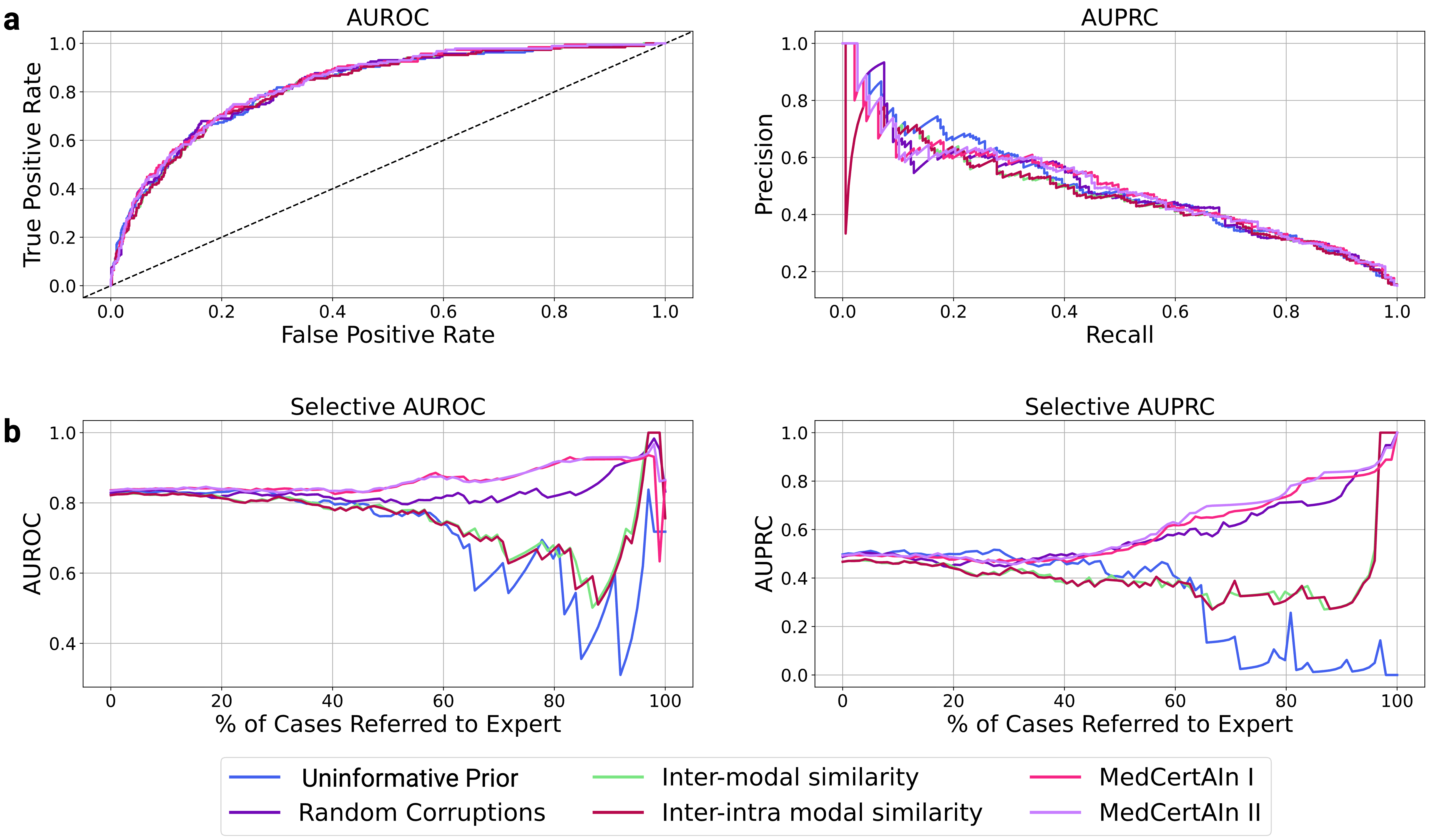}
    \caption{\small
    \textbf{Comparison Across Multimodal Data-Driven Priors.}
    Standard AUROC and AUPRC curves show modest differences across prior constructions, whereas selective AUROC and AUPRC reveal clearer separation across deferral rates. Combining similarity-based selection with random corruptions in \texttt{MedCertAIn I} and \texttt{MedCertAIn II} yields the strongest selective performance.
    }
    \label{fig:figure_5}
\end{figure}

\Cref{fig:figure_5} visualizes the standard and selective prediction trends corresponding to the prior ablations in \Cref{table:table_6} as reported in the main text.

\clearpage  
\section{Derivation of Variational Objective}
\label{sec:variational_objective}

We adapt the approach of \citet{rudner2024uap} to the multimodal setting and provide the complete derivation verbatim below.

\subsection{A Family of Data-Driven Priors}
\label{appsec:data_driven_priors}

Consider a parametric observation model $p_{Y | X, \Theta}(y \vbar x, \theta; f)$, and let the mapping $f$ be defined by \mbox{$f(\cdot \,; \theta) \defines h(\cdot \,; \theta_{h}) \theta_{L}$}, where $h(\cdot \,; \theta_{h})$ is the post-activation output of the penultimate layer, $\Theta_{L}$ is the set of stochastic final-layer parameters, $\Theta_{h}$ is the set of stochastic non-final-layer parameters, and $\Theta \defines \{ \Theta_{h} , \Theta_{L}\}$ is the full set of stochastic parameters.
We assume access to pre-trained feature parameters, $\theta_{h}^{\ast}$, and context data that encodes useful information about the downstream tasks.
We denote a batch of context inputs with corresponding context labels by \mbox{$x_{\mathrm{c}} = \{ x_{1}, ..., x_{M} \}$} and \mbox{$y_{\mathrm{c}} = \{ y_{1}, ..., y_{M} \}$}, respectively, and let $p_{\smash{{X}_{c}, {Y}_{c}}}$ be a joint distribution over context batches.

To construct a family of data-driven priors, we begin by specifying a {\em context inference problem}.
We consider a Bernoulli random variable $\check{Z}$ denoting whether a given set of neural network parameters induces predictions that exhibit some desired property (e.g., high uncertainty on a set of evaluation points).
Furthermore, we define a {\em context observation model} $\check{p}_{\smash{\check{Z} \vbar \Theta}}(\check{z} \vbar \theta ; f, p_{\smash{{X}_{c}, {Y}_{c}}})$---which denotes the likelihood of observing a yet-to-be-specified outcome $\check{z}$ under $\check{p}_{\smash{\check{Z} \vbar \Theta}}$ given $\theta$ and $p_{\smash{{X}_{c}, {Y}_{c}}}$---and specify a {\em base prior} over the model parameters, $p_{\Theta}(\theta)$.
For notational simplicity, we will drop the subscripts going forward except when needed for clarity.
With this setup, we can now define the context inference problem as finding the conditional distribution over neural network parameters that we {\it would} obtain if we conditioned on the desired property being satisfied.
This conditional distribution will serve as our data-driven prior, and by Bayes' Theorem, we can express it as
\begin{align}
    p(\theta \vbar \check{z} ; p_{\smash{{X}_{c}, {Y}_{c}}})
    =
    \frac{ \check{p}(\check{z} \vbar \theta ; p_{\smash{{X}_{c}, {Y}_{c}}}) p(\theta) }{ \check{p}(\check{z} ; p_{\smash{{X}_{c}, {Y}_{c}}}) } .
    \label{eq:data-driven_prior}
\end{align}
To define a family of data-driven priors that place high probability density on neural network parameter values that induce predictive functions with reliable uncertainty estimates, we specify a Bernoulli context observation model $\check{p}_{\smash{\check{Z} \vbar \Theta}}$ in which $\check{Z} = 1$ denotes the outcome of `achieving reliable uncertainty quantification' and $\check{p}(\check{z} = 1 \vbar \theta ; p_{\smash{{X}_{c}, {Y}_{c}}})$ denotes the likelihood of $\check{z} = 1$ given $\theta$ and $p_{\smash{{X}_{c}, {Y}_{c}}}$.
More specifically, we define:
\begin{align}
\begin{split}
\label{eq:aux_likelihood}
    \check{p}(\check{z} = 1 \vbar \theta ; p_{\smash{{X}_{c}, {Y}_{c}}})
    &
    =
    \exp(- \mathbb{E}_{p_{\smash{{X}_{c}, {Y}_{c}}}}[c({X}_{c}, {Y}_{c}, \theta) ] )
    \\[10pt]
    \check{p}(\check{z} = 0 \vbar \theta ; p_{\smash{{X}_{c}, {Y}_{c}}})
    &
    =
    1 - \check{p}(\check{z} = 1 \vbar \theta ; p_{\smash{{X}_{c}, {Y}_{c}}}) ,
\end{split}
\end{align}
where $c : \mathcal{X} \times \mathcal{Y} \times \mathbb{R}^{P} \rightarrow \mathbb{R}_{\geq}$ is a {\em cost function}.
By specifying the outcome $\check{z} = 1$ along with a distribution, $p_{\smash{{X}_{c}, {Y}_{c}}}$ we obtain a conditional distribution $\check{p}(\theta \vbar \check{z} ; f, p_{\smash{{X}_{c}, {Y}_{c}}})$---the distribution over neural network parameters that we {\em would} infer if we observed the outcome $\check{z} = 1$ under the base prior and the Bernoulli context observation model defined in~\Cref{eq:aux_likelihood}.
Naturally, the quality (i.e., the usefulness) of this conditional distribution is determined by the quality of the context observation model $\check{p}_{\smash{\check{Z} \vbar \Theta}}$, the data, and the prior.
As a result, the primary challenge in designing effective uncertainty-aware priors lies in constructing a context observation model---via a cost function $c$ and a context distribution $p_{\smash{{X}_{c}, {Y}_{c}}}$---that is as well-specified as possible.
The better specified the context observation model, the more \mbox{useful the data-driven prior.}

\subsection{Data-Driven, Uncertainty-Aware Priors for Fine-Tuning Pre-trained Models}

In this section, we present a specific instantiation of an uncertainty-aware prior for fine-tuning pre-trained models.
To define a data-driven prior $\check{p}(\theta \vbar \check{z} ; p_{\smash{{X}_{c}, {Y}_{c}}})$ that incorporates useful information from the pre-trained parameters $\theta_{h}^{\ast}$ and assigns high probability density to parameter values $\theta$ that induce models with reliable uncertainty quantification, we need to specify a suitable context likelihood and suitable layer-specific base priors $p(\theta_{h})$ and $p(\theta_{L})$.
For the base priors, we let $\smash{p(\theta_{h}) = \calN(\theta_{h} ; \theta_{h}^{\ast}, \tau_{h}^{-1} I)}$, which assigns high probability to parameters $\theta_{h}$ that are close to the pre-trained parameters $\theta_{h}^{\ast}$, and $p(\theta_{L}) = \calN(\theta_{L} ; \mathbf{0}, \tau_{L}^{-1} I)$.

To define a context observation model that induces a data-driven prior with desirable properties, we specify a cost function $c$ of the form
\begin{align}
\begin{split}
\label{eq:regularizer}
    c({x}_{c}, {y}_{c}, \theta)
    \defines
    \tau\hspace*{-1.5pt} \sum_{k=1}^{K} D^{2}_{\mathcal{M}}([f({x}_{c} ; \theta)]_{k}, m({x}_{c}, {y}_{c})_{k}, {C}({x}_{c})),
\end{split}
\end{align}
where $K$ is the number of output dimensions, \mbox{$\smash{p_{{X}_{c}, {Y}_{c}}}$} is a joint distribution over context batches ${x}_{c}$ and ${y}_{c}$ (each of size $M$),
\begin{align}
\label{eq:mahalanobis_distance}
    D^{2}_{\mathcal{M}}([f({x}_{c} ; \theta)]_{k}, m({x}_{c}, {y}_{c})_{k}, {C}({x}_{c}))
    \defines
    \mathbf{v}_{k}^\top {C}({x}_{c})^{-1} \mathbf{v}_{k}
\end{align}
with $\mathbf{v}_{k} \defines [f({x}_{c} ; \theta)]_{k} - m({x}_{c}, {y}_{c})_{k}]$ is the squared Mahalanobis distance between model predictions $[f({x}_{c} ; \theta)]_{k}$ and an input-dependent distribution with mean $m({x}_{c}, {y}_{c})_{k}$ and $M$-by-$M$ covariance matrix ${C}({x}_{c})$.
To obtain a data-driven prior that assigns high probability density to parameters $\theta$ that induce models with reliable uncertainty estimates, we specify a data-dependent mean function, $m({x}_{c}, {y}_{c})_{k} \defines [y_{c}]_{k}$, and a covariance function
\begin{align}
    C(\cdot)
    \defines
     s_{1} h(\cdot ; \theta^{\ast}_{h}) h(\cdot ; \theta^{\ast}_{h})^\top + s_{2} I ,
    \label{eq:covariance}
\end{align}
parameterized by pre-trained model parameters $\theta^{\ast}_{h}$ and fixed scaling parameters $\tau$, $s_{1}$, and $s_{2}$, that reflects structure in the pre-trained model representations $h(\cdot ; \theta^{\ast}_{h})$.
Finally, we define the context distribution as $p(x_{\mathrm{c}}, y_{\mathrm{c}}) = p(y_{\mathrm{c}} \vbar x_{\mathrm{c}}) p(x_{\mathrm{c}})$, where $$p(y_{\mathrm{c}} \vbar x_{\mathrm{c}}) \defines \delta(\{\mathbf{0}, ..., \mathbf{0} \} - y_{\mathrm{c}})$$ and $p(x_{\mathrm{c}})$ is an empirical distribution constructed from a larger set of (domain- and task-specific) context inputs.\footnote{Defining $p(y_{\mathrm{c}} \vbar x_{\mathrm{c}}) \defines \delta(\{\mathbf{0}, ..., \mathbf{0} \} - y_{\mathrm{c}})$ implies that under $p_{\smash{{X}_{c}, {Y}_{c}}}$, all context batch samples have $y_{c} = \mathbf{0}$, and therefore, we effectively have $m({x}_{c}, {y}_{c})_{k} \defines \mathbf{0}$ for all context batch samples.}

Under this cost function and context distribution, the data-driven prior defined in \Cref{eq:aux_likelihood}, by design, assigns high probability density to parameters $\theta$ that induce predictions $f({x}_{c} ; \theta)$ that have high predictive uncertainty on the context inputs.
If the distribution over context inputs, $p_{\smash{{X}_{c}}}$, is specified to place high probability density on context batches which contain input points that are meaningfully distinct from the training inputs, then the data-driven prior favours models that exhibit high predictive uncertainty on such meaningfully distinct inputs.

\subsection{Variational Inference with Uncertainty-Aware Priors}

In this section, we show how to perform variational inference with uncertainty-aware priors.
We start by specifying a probabilistic model with an uncertainty-aware prior,
\begin{align}
\begin{split}
    p(y_{\calD}, \theta \vbar x_{\calD}, \check{z} ; p_{\smash{{X}_{c}, {Y}_{c}}})
    =
    \underbrace{p(y_{\calD} \vbar x_{\calD}, \theta ; f)}_{\textrm{Likelihood}}
    \hspace*{7pt}
    \cdot
    \underbrace{p(\theta \vbar \check{z} ; p_{\smash{{X}_{c}, {Y}_{c}}})}_{\textrm{Uncertainty-aware prior}}\hspace*{-7pt}
    .
\end{split}
\end{align}
To perform variational inference in this model and approximate the posterior distribution over the parameters of interest, we begin by defining a variational distribution,
\begin{align}
    q(\theta)
    \defines
    q(\theta_{h}) \, q(\theta_{L}) ,
\end{align}
where \mbox{$q(\theta_{L}) = \calN(\theta_{L} ; \mu_{L}, \Sigma_{L})$} and \mbox{$q(\theta_{h}) = \calN(\theta_{h} ; \mu_{h}, \Sigma_{h})$} with learnable variational parameters $\mu \defines \{ \mu_{h}, \mu_{L} \}$ and $ \Sigma \defines \{ \Sigma_{h}, \Sigma_{L} \}$, and frame the inference problem of finding the posterior $p(\theta \vbar x_{\calD}, y_{\calD}, \check{z})$ variationally as
\begin{align}
    \min_{q_{\Theta} \in \calQ} \DKL{q_{\Theta}}{p_{\Theta \vbar X_{\calD}, Y_{\calD}, \check{Z}}} ,
\end{align}
where $\smash{\calQ}$ is a mean-field Gaussian variational family.
This variational problem can equivalently be expressed as maximizing the variational objective
\begin{align}
\begin{split}
    \bar{\calF}(q_{\Theta})
    \defines
    \mathbb{E}_{q_{\Theta}} [ \log p(y_{\calD} \vbar x_{\calD} , \Theta ; f) ]
    - \DKL{q_{\Theta}}{p_{\Theta \vbar \smash{\check{Z}}}} .
\end{split}
\end{align}
Unfortunately, this variational objective is intractable since the data-driven prior $\check{p}(\theta \vbar \check{z} ; p_{\smash{{X}_{c}, {Y}_{c}}})$ defined in \Cref{eq:data-driven_prior}---which is required to compute $\DKL{q_{\Theta}}{p_{\Theta \vbar \check{Z}}}$---is not in general tractable.

To overcome this intractability, we take advantage of the properties of the \kld and note that we can express $\DKL{q_{\Theta}}{p_{\Theta \vbar \check{Z}}}$ as:
\begin{align}
\begin{split}
    \DKL{q_{\Theta}}{p_{\Theta \vbar \smash{\check{Z}}}}
    &
    =
    \mathbb{E}_{q_{\Theta_{h}}q_{\Theta_{L}}} \left[\log \frac{q(\Theta_{h}) \, q(\Theta_{L})}{p(\Theta_{h}) \, p(\Theta_{L}) } \right]
    - \mathbb{E}_{q_{\Theta_{h}}q_{\Theta_{L}}} [\log \check{p}(\check{z} \vbar \Theta ; p_{\smash{{X}_{c}, {Y}_{c}}}) ]
    + \log \check{p}(\check{z} ; p_{\smash{{X}_{c}, {Y}_{c}}}) ,
\end{split}
\end{align}
where the intractable log-marginal likelihood $\smash{\log \check{p}(\check{z} ; p_{\smash{{X}_{c}, {Y}_{c}}})}$ was factored out as an additive constant independent of any learnable parameters.
Using this result, we can obtain a tractable upper bound on the KL divergence:
\begin{align}
\begin{split}
    \log \check{p}(\check{z} ; p_{\smash{{X}_{c}, {Y}_{c}}})
    &
    \geq
    \mathbb{E}_{q_{\Theta_{h}}q_{\Theta_{L}}} [\log \check{p}(\check{z} \vbar \Theta ; p_{\smash{{X}_{c}, {Y}_{c}}}) ]
    - \DKL{q_{\Theta_{h}}}{p_{\Theta_{h}}}
    - \DKL{q_{\Theta_{L}}}{p_{\Theta_{L}}} ,
\end{split}
\end{align}
where each KL divergence term can be computed analytically, and we can obtain an unbiased estimator of the negative log-likelihood using simple Monte Carlo estimation.
Using this inequality, we can obtain a lower bound on the log marginal likelihood and obtain a valid variational objective.


\paragraph{Variational Objective.}
Since $q_{\Theta_{h}}$ and $q_{\Theta_{L}}$ are both mean-field Gaussian distributions, we can obtain a doubly lower bounded variational objective
\begin{align}
\begin{split}
    \calF(\mu, \Sigma)
    &
    \defines
    \underbrace{\mathbb{E}_{q_{\Theta}} [ \log p(y_{\calD} \vbar x_{\calD} , \Theta ; f) ]}_{\textrm{Expected log-likelihood}}
    - \hspace*{-3pt}\underbrace{\DKL{q_{\Theta_{h}}}{p_{\Theta_{h}}}}_{\textrm{Pre-training regularization}} \hspace*{-3pt} - \hspace*{0pt}\underbrace{\DKL{q_{\Theta_{L}}}{p_{\Theta_{L}}}}_{\textrm{Final-layer regularization}}
    - \underbrace{\mathbb{E}_{q_{\Theta}}[ \mathbb{E}_{p_{\smash{{X}_{c}, {Y}_{c}}}}[c({X}_{c}, {Y}_{c}, \Theta) ] ]}_{\textrm{Uncertainty regularization}}
    ,
    \label{eq:final_objective}
\end{split}
\end{align}
where the cost function and context distribution are as defined above.
We can estimate all expectations in the objective using simple Monte Carlo estimation, giving the final variational objective:
\begin{align}
\begin{split}
    \hat{\calF}(\mu, \Sigma)
    &
    \defines
    \frac{1}{J}\sum_{j=1}^{J} \log p(y_{\calD} \vbar x_{\calD} , \theta^{(j)} ; f)
    - \DKL{q_{\Theta}}{p_{\Theta}}
    - \frac{1}{J J'}\sum\nolimits_{j=1}^{J} \sum\nolimits_{j'=1}^{J'} c({x}_{c}^{(j')}, {y}_{c}^{(j')}, \theta^{(j)})
    ,
    \label{eq:final_mc_objective}
\end{split}
\end{align}
with $\theta^{(j)} \sim q_{\Theta}$, ${x}_{c}^{(j')} \sim p_{X_{c}}$, and ${y}_{c}^{(j')} \sim p_{Y_{c} | X_{c}}$ for $j = 1, ..., J$ and $j' = 1, ..., J'$.
This objective can be maximized with stochastic variational inference~\citep{hoffman2013svi}.

\clearpage

\section{Experimental Details}
\label{app:experimental-details}

\subsection{Dataset and Training Details}

\setlength{\tabcolsep}{5pt}
\begin{table}[h]
\caption{\small \textbf{Task-Specific Dataset Sizes.} Numbers of unimodal and paired multimodal samples derived from MIMIC-IV (EHR) and MIMIC-CXR (CXR) \citep{mimiccxr,mimiciv} for in-hospital mortality prediction. The multimodal cohort is smaller because it includes only samples for which both modalities are available.}
\small
\label{table:dataset_sizes}
\vspace*{5pt}
\centering
\begin{tabular}{lccc}
\toprule
\textbf{Dataset} & \textbf{Training} & \textbf{Validation} & \textbf{Testing} \\
\midrule
Unimodal CXR (images) & 124,671 & 8,813 & 20,747 \\
Unimodal EHR (time-series records) & 42,628 & 4,802 & 11,914 \\
Multimodal (CXR + EHR) & 4,485 & 488 & 1,242 \\
\bottomrule
\end{tabular}
\end{table}

\subsection{ConVIRT Training Details}
\label{appsec:convirt_training}

Contrastive VIsual Representation Learning from Text (ConVIRT) \citep{zhang2022contrastive} is a self-supervised framework originally developed to learn representations from paired medical images and radiology reports.
We adapt its bidirectional contrastive objective to paired EHR and CXR data, encouraging matched multimodal representations to align relative to unmatched pairs.
The loss for our ConVIRT-MedFuse architecture is
$\mathcal{L}=\mathcal{S}(x^{\textrm{ehr}},x^{\textrm{cxr}})+\mathcal{S}(x^{\textrm{cxr}},x^{\textrm{ehr}})$,
where $\mathcal{S}$ denotes the InfoNCE loss.

We performed 10 random-search trials, sampling the learning rate from a uniform distribution over $[10^{-2},10^{-1}]$.
We used a batch size of 256 and trained each configuration for 300 epochs without early stopping, and selected the checkpoint with the highest validation AUROC across all epochs and configurations.
Using the selected model, we performed inference on the original paired training data $(X^{\textrm{ehr}},X^{\textrm{cxr}})$ to obtain latent representations and compute cross-modal cosine similarity.

\subsection{Stochastic Hyperparameter Search Space}
\label{appsec:stochastic_hyper_tuning}

\setlength{\tabcolsep}{5pt}
\begin{table}[h]
\caption{\textbf{Stochastic Hyperparameter Search Space.} Hyperparameters and candidate values evaluated for the uncertainty-aware prior in \texttt{MedCertAIn}.}
\vspace*{5pt}
\small
\label{table:stochastic_hyperparams}
\centering
\begin{tabular}{ll}
\toprule
\textbf{Hyperparameter} & \textbf{Values} \\
\midrule
Prior variance & ${0.1,1,10,1000}$ \\
Prior likelihood scale & ${0.1,1,10}$ \\
Prior likelihood f-scale & ${0,1,10}$ \\
Prior likelihood covariance scale & ${0.1,0.01,0.001}$ \\
Prior likelihood covariance diagonal & ${0.5,1,5}$ \\
\bottomrule
\end{tabular}
\end{table}

\subsection{Technical Implementation}

The data-loading and preprocessing pipeline was implemented in PyTorch \citep{pytorch}, following the structure of the original MedFuse implementation \citep{medfuse}.
We refactored the unimodal and multimodal models and their training and evaluation loops in JAX \citep{jax}.
JAX facilitated the implementation of the Bayesian neural networks and stochastic training procedures underlying the uncertainty quantification methods evaluated in this work.
In our experiments, this refactoring also reduced the total training time relative to the original PyTorch implementation.

\end{document}